\pdfoutput=1

\documentclass[11pt,a4paper]{article}
\usepackage[hyperref]{acl2021}
\usepackage{times}
\usepackage{latexsym}

\usepackage{microtype}

\usepackage{multirow}
\usepackage{mathtools, nccmath}
\usepackage{siunitx}
\usepackage{commath}
\usepackage{amsmath}
\usepackage{enumitem}
\setlist[enumerate]{nosep,topsep=1pt}
\setlist[itemize]{nosep,topsep=1pt}
\usepackage{tabularx}
\newcolumntype{Y}{>{\centering\arraybackslash}X}
\newcolumntype{R}[1]{>{\raggedleft\arraybackslash}p{#1}}
\newcolumntype{L}[1]{>{\raggedright\arraybackslash}p{#1}}
\newcolumntype{P}[1]{>{\centering\arraybackslash}p{#1}}
\usepackage{pgfplots}
\usepackage{pgfplotstable}
\usetikzlibrary{patterns}
\usetikzlibrary{intersections}
\usetikzlibrary{calc}
\usepackage{subcaption}
\usepackage[ruled,vlined,noend]{algorithm2e}
\usepgfplotslibrary{colorbrewer}
\usepackage{booktabs}
\renewcommand{\paragraph}{%
  \@startsection{paragraph}{4}%
  {\z@}{0ex \@plus .2ex \@minus .2ex}{-1em}%
  {\normalfont\normalsize\bfseries}%
}
\usepackage{makecell}
\usepackage[]{todonotes}

\aclfinalcopy 
\def\aclpaperid{1772} 

\title{Societal Biases in Language Generation: Progress and Challenges}

\author{Emily Sheng$^1$, Kai-Wei Chang$^2$,  Premkumar Natarajan$^1$,  Nanyun Peng$^{1,2}$ \\
 $^1$ Information Sciences Institute, University of Southern California \\
 $^2$ Computer Science Department, University of California, Los Angeles \\
 {\tt \{ewsheng,pnataraj\}@isi.edu},  {\tt \{kwchang,violetpeng\}@cs.ucla.edu} \\}

\date{}

\begin{document}
\maketitle
\begin{abstract}
 Technology for language generation has advanced rapidly, spurred by advancements in pre-training large models on massive amounts of data and the need for intelligent agents to communicate in a natural manner.
 While techniques can effectively generate fluent text, they can also produce undesirable societal biases that can have a disproportionately negative impact on marginalized populations.
 Language generation presents unique challenges for biases in terms of direct user interaction and the structure of decoding techniques.
 To better understand these challenges, we present a survey on societal biases in language generation, focusing on how data and techniques contribute to biases and progress towards reducing biases.
 Motivated by a lack of studies on biases from decoding techniques, we also conduct experiments to quantify the effects of these techniques.
 By further discussing general trends and open challenges, we call to attention promising directions for research and the importance of fairness and inclusivity considerations for language generation applications.
\end{abstract}

\section{Introduction}
Natural language generation (NLG) is a suite of techniques that enables the generation of human-readable language for different goals.
These techniques are the core components of applications such as virtual assistants, chat bots, automatic translators, summarizers, and creative language composers.
Recent advances in techniques for language generation (e.g., GPT \citep{radford2018improving}, GPT-2 \citep{radford2019language}, GPT-3 \citep{brown2020language}, TransformerXL \citep{dai-etal-2019-transformer}, XLNet \citep{yang2019xlnet}) powered by Transformers \citep{vaswani2017attention} and an increasing repository of available data have created more capable applications.
This has, in turn, channeled more interest and effort into developing NLG techniques.

We emphasize the importance of better understanding how societal biases manifest in NLG techniques, because NLG applications \emph{directly interact} with many different users to generate novel content in various domains (e.g., chat bots for health, education, and customer support).
However, when techniques are less effective or detrimental for marginalized populations, these techniques can inadvertently become \emph{gatekeepers} of those populations for generation and associated language technologies.
For example, an educational chat bot that produces more negative responses for topics about a specific ethnicity will discourage users of that ethnicity from interacting with the chat bot.
While it is generally important to study the societal impact of NLP and AI techniques, we argue that the direct user impact of NLG techniques makes it especially important to carefully quantify the impact.

\begin{table*}[!t]{
\footnotesize
\centering
    \begin{tabularx}{\linewidth}{L{5.5em} L{5em} X}
    \hline
    \toprule
    \textbf{Demo. Dim.} & \textbf{NLG Task} & \textbf{Works} \\ \midrule
    \textbf{Gender} & Autocomplete & \citet{bordia-bowman-2019-identifying,qian-etal-2019-reducing,solaiman2019release,sheng2019woman,sheng-etal-2020-towards,vig2020investigating,yeo-chen-2020-defining,brown2020language,dhamala2021bold,schick2021self,nozza2021honest,kirk2021true}
    \\
    & Dialogue & \citet{henderson2018ethical,dinan-etal-2020-queens,liu-etal-2020-gender,liu-etal-2020-mitigating,cercas-curry-etal-2020-conversational,sheng2021revealing,sheng2020nice} \\ 
    & MT & \citet{vanmassenhove-etal-2018-getting,elaraby2018gender,prates2019assessing,stanovsky-etal-2019-evaluating,escude-font-costa-jussa-2019-equalizing,cho-etal-2019-measuring,moryossef-etal-2019-filling,saunders-byrne-2020-reducing,saunders-etal-2020-neural,kocmi-etal-2020-gender,costa-jussa-de-jorge-2020-fine,costa2020gender,basta-etal-2020-towards,farkas2020measure,stafanovics-etal-2020-mitigating,gonen-webster-2020-automatically,hovy-etal-2020-sound,roberts2020decoding,cho2021towards,savoldi2021gender,renduchintala2021investigating,choubey2021improving,saunders2021first,tomalin2021practical} \\ 
    & Re-writing & \citet{habash-etal-2019-automatic,zmigrod-etal-2019-counterfactual,alhafni-etal-2020-gender,sun2021they} \\ \cmidrule{1-3}
    
    \textbf{Profession} & Autocomplete & \citet{huang-etal-2020-reducing,dhamala2021bold} \\ \cmidrule{1-3}
    
    \textbf{Race} & Autocomplete & \citet{solaiman2019release,sheng2019woman,sheng-etal-2020-towards,groenwold-etal-2020-investigating,brown2020language,dhamala2021bold,schick2021self,kirk2021true} \\
    & Dialogue & \citet{sheng2021revealing,sheng2020nice} \\ \cmidrule{1-3}
    
    \textbf{Religion} & Autocomplete & \citet{solaiman2019release,brown2020language,dhamala2021bold,kirk2021true,abid2021persistent} \\ \cmidrule{1-3}
    
    \textbf{Sexuality} & Autocomplete & \citet{sheng2019woman,sheng-etal-2020-towards,kirk2021true} \\
    & Dialogue & \citet{sheng2021revealing} \\
    \cmidrule{1-3}
    
    \textbf{Other} & Autocomplete & \citet{shwartz-etal-2020-grounded,peng-etal-2020-reducing,huang-etal-2020-reducing,dhamala2021bold,kirk2021true} \\
    & Dialogue & \citet{sheng2021revealing} \\
    & Re-writing & \citet{pryzant2020automatically,ma-etal-2020-powertransformer} \\
    \bottomrule
    \end{tabularx}
}
\vspace{-0.5em}
\caption{\label{tab:group-fairness} Existing bias studies on different demographic dimensions in various NLG tasks: autocomplete generation, dialogue generation, machine translation (MT), and text re-writing.}
\vspace{-1.5em}
\end{table*}

Motivated by the importance of fairness in language generation, we present the first comprehensive survey on societal biases in language generation.
By enumerating how NLG techniques contribute to biases and examining progress towards bias analysis and mitigation, we contextualize the discussion of broader trends and challenges.
Specifically, we focus on techniques for NLG tasks, i.e., tasks that generate a sequence of text.\footnote{\label{footnote:lm}Although bi-directional language models like BERT \citep{devlin-etal-2019-bert} can also be used for auto-regressive generation \citep{wang-cho-2019-bert,chen-etal-2020-distilling}, traditional auto-regressive models are still typically of better quality and more widely used for generation \citep{shwartz-etal-2020-grounded}. Thus, we limit the scope of this survey to the latter models.}
Finding a lack of studies on biases from decoding techniques, we additionally present an experimental study to quantify the effects of various decoding techniques. 

Before we delve into the details of biases in language generation, we first position our survey in the context of other relevant surveys and position papers.
\citet{sun-etal-2019-mitigating} present a focused survey on mitigating gender biases and \citet{shah-etal-2020-predictive} categorize sources of biases---both largely focus on natural language understanding (NLU) tasks, while we examine biases in NLG tasks.
Additionally, \citet{blodgett-etal-2020-language} urge for more explicitly tying ``biases'' in NLP to societal normative definitions of biases and social hierarchies; with their recommendations in mind, we discuss the negative impacts of biases in NLG techniques.

Our contributions are a comprehensive survey on societal biases in language generation and an experimental study on biases from decoding techniques.
To start, we describe classes of NLG tasks (Sec.~\ref{sec:tasks}) and subsequently examine examples of biases and harms in NLG (Sec.~\ref{sec:biases-harms}).
We then discuss NLG techniques that facilitate biases, including a study of decoding techniques (Sec.~\ref{sec:characteristics}).
Sec.~\ref{sec:progress} highlights progress and challenges, and Sec.~\ref{sec:open} presents open problems and proposals.
We hope this survey brings more visibility to the importance of carefully considering different components of NLG pipelines for potential biases and mitigation methods.

\section{Language Generation Tasks} 
\label{sec:tasks}
To begin, we categorize generation tasks and introduce existing bias studies relevant to each task.
NLG tasks broadly fall into two categories: those that \emph{generate text continuations} conditioned on some prompt and those that \emph{transform text from one form to another}.
Table~\ref{tab:group-fairness} organizes various bias-related works for NLG tasks.

\subsection{Continuation Generation Tasks}
The continuation class includes autocomplete and dialogue generation, where the goal is to generate text that is coherent and relevant to a prompt. 

\paragraph{Autocomplete Generation}
We use the term autocomplete generation to refer to conditional generation directly from language models.
Language models are the core components for many NLG and NLU tasks, and this task enables directly quantifying biases in large, pre-trained language models \citep{bordia-bowman-2019-identifying,sheng2019woman,solaiman2019release,brown2020language}.
Existing works analyzing biases in autocomplete generation have mostly examined Transformer-based models, including GPT \citep{shwartz-etal-2020-grounded}, GPT-2 \citep{solaiman2019release,sheng2019woman,sheng-etal-2020-towards,shwartz-etal-2020-grounded,vig2020investigating,yeo-chen-2020-defining,huang-etal-2020-reducing,dhamala2021bold,schick2021self}, GPT-3 \citep{brown2020language}, CTRL \citep{dhamala2021bold}, TransformerXL \citep{shwartz-etal-2020-grounded,vig2020investigating,huang-etal-2020-reducing}, and XLNet \citep{shwartz-etal-2020-grounded,vig2020investigating,yeo-chen-2020-defining}, though \citet{bordia-bowman-2019-identifying,qian-etal-2019-reducing} also look at LSTM-based models.

\paragraph{Dialogue Generation}
Dialogue generation is conditioned on user inputs and can be for specific domains (e.g., health, customer service) and tasks (e.g., behavior intervention, booking flights) or general chit-chat.
These dialogue applications directly interact with users, and any propagated biases directly affect user behavior and actions.
In terms of recurrent dialogue models, \citet{henderson2018ethical} analyze biases in hierarchical recurrent encoder-decoder architectures
and \citet{liu-etal-2020-gender,liu-etal-2020-mitigating} analyze LSTM-based encoder-decoder models.
Other works on dialogue biases \citep{dinan-etal-2020-queens,sheng-etal-2020-towards,sheng2020nice} focus on Transformer-based models such as DialoGPT \citep{zhang-etal-2020-dialogpt} and other custom architectures.

\subsection{Transformation Generation Tasks}
The transformation class includes machine translation and various formulations of text re-writing.
The general goal of these tasks is to transform text into a form with targeted properties.

\paragraph{Machine Translation}
Translation is the task of transforming text between languages while preserving the meaning.
Existing works on biases in machine translation have almost exclusively focused on issues of gender biases\footnote{
For a detailed survey of gender bias in machine translation, we refer readers to \citet{savoldi2021gender}.} in a variety of academic and commercial systems.
The use of grammatical gender in some languages and not in others can expose unwanted gender associations (e.g., for different occupations) through translation \citep{prates2019assessing}.
Earlier works by \citet{vanmassenhove-etal-2018-getting} and \citet{elaraby2018gender} study LSTM-based encoder-decoder translation systems, and more recent works examine 
Transformer-based architectures \citep{escude-font-costa-jussa-2019-equalizing,stanovsky-etal-2019-evaluating,saunders-byrne-2020-reducing,saunders-etal-2020-neural,costa-jussa-de-jorge-2020-fine,basta-etal-2020-towards,stafanovics-etal-2020-mitigating,renduchintala2021investigating,choubey2021improving,saunders2021first,tomalin2021practical}.
While Google Translate\footnote{\url{https://translate.google.com}} has been the most popular commercial system to analyze for gender biases \citep{prates2019assessing,moryossef-etal-2019-filling,stanovsky-etal-2019-evaluating,cho-etal-2019-measuring,farkas2020measure}, \citet{stanovsky-etal-2019-evaluating} also study Microsoft Translator,\footnote{\url{https://www.bing.com/translator}} Amazon Translate,\footnote{\url{https://aws.amazon.com/translate}} and SYSTRAN;\footnote{\url{https://www.systransoft.com}} \citet{cho-etal-2019-measuring} additionally look at Naver Papago\footnote{\url{https://papago.naver.com}} and Kakao Translator,\footnote{\url{https://translate.kakao.com}} and \citet{cho2021towards} also examine Yandex.\footnote{\url{https://translate.yandex.com}}

\paragraph{Re-writing}
We use the term re-writing to refer to tasks of revising specific words and phrases in the original text to be more aligned with a targeted attribute.
Specifically, there have been studies on re-inflection \citep{habash-etal-2019-automatic,zmigrod-etal-2019-counterfactual,alhafni-etal-2020-gender} and re-writing text to use neutral viewpoints \citep{pryzant2020automatically}, gender-neutral English \citep{sun2021they},
or more agency \citep{ma-etal-2020-powertransformer}.
These tasks typically rely on custom encoder-decoder models.

\subsection{Other Tasks}
There are other NLG tasks, such as the continuation tasks of story and poetry generation, and the transformation tasks of abstractive summarization and paraphrase generation.
However, these other NLG tasks are not yet well-studied in the context of societal biases.\footnote{\citet{lucy2021gender} is an exception that analyzes gender in generated stories. While there are studies of biases in poetry generation and summarization, they focus on non-NLG biases: \citet{sheng-uthus-2020-investigating} investigate biases in a poetry composition system, but in the context of information retrieval; \citet{celis2020dialect} analyze biases in extractive summarization.}

\section{Biases and their Negative Impacts}
\label{sec:biases-harms}
In this section, we introduce how existing studies of biases in NLG tasks commonly quantify biases and their negative impacts.

\subsection{Bias Definitions and Metrics}
In the context of AI fairness, the term ``bias'' commonly refers to skews that result in undesirable impacts \citep{crawford2017bias} and is quantifiable with some metric.
There are relatively more existing studies on biases in NLU tasks, where it is arguably simpler to define bias metrics, since we can intuitively compare the accuracy of the task (e.g., coreference resolution, hate speech detection) for different demographics.
Language generation tasks often involve stochastic generation of open-ended and lengthy texts, traits that are not directly compatible with traditional algorithmic bias definitions (e.g., equalized odds, equal opportunity, demographic parity \citep{dwork2012fairness,hardt2016equality}).

Because of the difficulty in defining metrics, existing works define bias loosely as demographic inequality and use intermediate proxy metrics to comparatively measure bias.
Examples include:
\begin{itemize}[leftmargin=*]
     \item \textbf{Regard Ratio}: \textit{negative-neutral-positive} regard score ratios of text generated from bias-inducing prompts \citep{sheng2019woman}
    \item \textbf{Sentiment Ratio}: \textit{negative-neutral-positive} sentiment score ratios of text generated from African American English (AAE) versus White-Aligned English (WAE) prompts \citep{groenwold-etal-2020-investigating} 
    \item \textbf{Individual and Group Fairness through Sentiment}: comparisons of the sentiment distributions of generated text across demographics and prompts \citep{huang-etal-2020-reducing}
    \item \textbf{Gendered Word Co-occurrence Score}: mean and standard deviations of the absolute log ratio of probabilities:
    $\mathcal{P}(\text{word}|\text{female terms})$ to $\mathcal{P}(\text{word}|\text{male terms})$ across all words in generated text \citep{bordia-bowman-2019-identifying}
\end{itemize}
There are also metrics for other bias evaluation setups in continuation generation tasks involving sentiment \citep{shwartz-etal-2020-grounded}, the ratio of gendered words \citep{solaiman2019release,vig2020investigating,dinan-etal-2020-queens}, and other novel metrics \citep{peng-etal-2020-reducing,yeo-chen-2020-defining}.
Studies of biases in transformation generation tasks favor metrics of accuracy in terms of successfully transforming text to have a desired property.
We present a more thorough comparison of metrics in Section~\ref{ssec:progress-evaluation}.

Bias metrics can also be categorized by how they define associations between demographic group attributes and text.
Biases can be towards people described in text, people who produce the text, or people to whom the text is addressed \citep{dinan-etal-2020-multi}.
Most existing works define bias metrics through the first association---these biases are relatively easier to analyze, since both the demographic and the textual signals of bias are encapsulated within the text.
There are also works that define biases towards people who produce the text
\citep{groenwold-etal-2020-investigating} or people to whom the text is addressed \citep{sheng2020nice}, though there are relatively fewer works that study these latter associations.

\subsection{Negative Impacts}
Biases in NLG techniques are important to study because they can result in harmful, negative impacts.
We survey detrimental representational\footnote{Unfair representations of different groups} and allocational\footnote{Unfair allocation of resources} impacts \citep{crawford2017bias,barocas2017problem,blodgett-etal-2020-language} used to motivate existing studies of bias in NLG tasks, finding limited examples.
While representational impacts are sometimes cited, it is difficult to measure the extent of the impacts.
Additionally, techniques for effective NLG are relatively new, and existing studies have limited knowledge of potential allocational impacts.
Finally, biases in NLG tasks give rise to a third type of negative impacts, which we call \textit{vulnerability impacts}.

\paragraph{Representational Impacts}
The works in Table~\ref{tab:group-fairness} motivate (to varying degrees) studying biases in NLG through potential negative representational impacts, in the form of propagating stereotypes, misrepresentations, or denigrations of social groups.
For example, \citet{sheng2019woman} enumerate how generated  text can propagate varying social perceptions of different demographics, 
and \citet{prates2019assessing} discuss how occupation-related gender biases could propagate stereotypes in translation.
However, it is difficult to quantify the effects of representational impacts;\footnote{\citet{kay2015unequal} is a rare example that explicitly studies the effect of representational impacts in image search.} while such impacts may be measured indirectly (e.g. by analyzing allocational impacts), we suggest long-term, interdisciplinary collaborations to explore the direct effects of these representational impacts.

\paragraph{Allocational Impacts} 
Harmful allocational impacts result from an unequal allocation of resources across groups.
Since effective NLG techniques based on large Transformer models \citep{vaswani2017attention} are relatively new,
most of the existing works on biases in NLG that list possible impacts only analyze direct representational consequences.
A real example of a negative allocational impact is when machine translation errors lead to arrests \citep{ong2017facebook}.
In general, technologies that are less effective or detrimental for certain populations become barriers that actively prevent those populations from using the technology, leading to diminished opportunities in jobs, education, health, etc.
We discuss more details in Section~\ref{ssec:deploying}.
With continuous technological advances, more organizations will turn to effective NLG techniques, making it imperative to start setting norms to reduce harmful allocational impacts \citep{tamkin2021understanding}.

\paragraph{Vulnerability Impacts}
Open-domain generation tasks can amplify a group's \emph{vulnerability to manipulation and harm}, which is an intermediate impact that makes a group more susceptible to representational and allocational impacts.
For example, privacy-related issues \citep{carlini2020extracting}, misinformation \citep{levy2021truth}, or radicalizing views in generated text could make a group more likely to be attributed to specific stereotypes (e.g., through action guided by misinformation) or end up with diminished opportunities (e.g., by having personal data exposed and misused).
Separately identifying vulnerability impacts could help facilitate recognition of other negative impacts.

\section{Contributors to NLG Biases} 
\label{sec:characteristics}
In a pipeline from data collection to evaluation for an NLG task, each component could propagate biases.\footnote{Task formulation and application deployment are also part of NLG task pipelines \citep{kiritchenko2020confronting}, though we do not focus on biases in these areas.}
We emphasize the ways in which data, model architecture, decoding, evaluation, and deployment uniquely exacerbate biases in generation tasks.
Additionally, we present an empirical study to show how measured biases in generated text can vary based on decoding technique.


\subsection{Biases from Data}
Modern NLP models often rely on large pre-trained language models, which in turn rely on a large collection of data to learn explicit and implicit associations.
Several recent pre-trained language models used for NLG tasks, e.g., T5 \citep{raffel2020exploring} and GPT-3 \citep{brown2020language}, are trained on the largest datasets used for any models.
These large models for generation are commonly trained on web data, which is known to contain biased language (e.g., \citet{ferrer2020discovering} discover gender, religion, and ethnic biases in Reddit communities).
While preprocessing is often included to filter out malformatted data and explicitly negative content (e.g., bad words and offensive phrases), those are generally the only efforts to reduce biases and associated impacts.
Furthermore, by filtering out all words deemed ``bad'', \citet{bender2021dangers} warns that we remove the discourse of marginalized populations.
\citet{paullada2020data}, \citet{bender-friedman-2018-data}, and \citet{gebru2018datasheets} provide more comprehensive surveys and frameworks that focus on aspects of data creation and management that could lead to biases, and we refer readers to their works for more discussion.
In the context of translation, \citet{cho2021towards} find that more data can increase translation fluency but may also make the system more biased.


\subsection{Biases from Model Architecture}
There are relatively few studies that examine model architectural properties that could lead to biases.
We discuss the few efforts towards understanding model biases in NLG tasks and emphasize the need for more to generalize.
For autocomplete generation, \citet{vig2020investigating} analyze GPT-2 variants through a causal mediation analysis, finding that larger models contain more gender bias, and bias tends to be  concentrated in a small number of neurons and attention heads.
\citet{silva2021towards} observe amplified biases in distilled versus original models.
For machine translation, \citet{costa2020gender} note that language-specific architectures are less biased because they encode more gender information than shared language encoder-decoder architectures.
Studies like the aforementioned are useful for designing targeted bias mitigation methods (e.g., controlled generation to target specific attention heads or regularization to retain gender information).
However, more evidence would be needed to generalize findings across models.\footnote{We also refer the reader to the work of \citet{park2018reducing} that discusses biases in NLU tasks from model components that ``attend'' to specific words (e.g., through attention or pooling), which could be applicable to NLG tasks as well.}

\subsection{Biases from Decoding}
\label{ssec:decoding}
While NLU and NLG models have structural similarities, NLG tasks uniquely use search or sampling techniques at inference time to generate text. Popular techniques include:
\begin{itemize}[leftmargin=*]
    \item \textbf{Greedy Search}: at each time step, choose the word with the highest probability.
    \item \textbf{Beam Search}: at each time step, keep the top $b$ hypotheses with highest probabilities; eventually pick the hypothesis with the highest probability.
    \item \textbf{Top-$k$ sampling} \citep{fan2018hierarchical}: at each time step, re-distribute the probability mass of the top $k$ words with highest probabilities and sample.
    \item \textbf{Nucleus sampling} \citep{holtzman2019curious}: at each time step, re-distribute the probability mass of the smallest set of words with a cumulative probability exceeding $p$ and sample.
\end{itemize}
More constrained forms of generation such as machine translation generally use variations of beam search; however, preferred decoding techniques are more varied for open-domain generation.
Despite variations in fluency and diversity between deterministic versus stochastic, search versus sampling procedures, there are limited studies \citep{roberts2020decoding} on how different decoding properties affect biases in generation.

\paragraph{A Study on Biases from Decoding}
To study how decoding techniques affect biases in generation, we use existing NLG bias metrics to evaluate text generated from different decoding methods.\footnote{Code at \url{https://github.com/ewsheng/decoding-biases}.}
We examine autocomplete generations from GPT, GPT-2, and XLNet, using the decoding techniques from Section~\ref{ssec:decoding}.
We evaluate with the following bias metrics: regard ratios \citep{sheng2019woman}, sentiment ratios \citep{groenwold-etal-2020-investigating}, individual and group fairness through sentiment scores \citep{huang-etal-2020-reducing}, and gendered word co-occurrence scores \citep{bordia-bowman-2019-identifying} (as introduced in Section~\ref{sec:biases-harms}).
More experimental details can be found in the Appendix.

In Section~\ref{ssec:progress-evaluation}, we distinguish between relative and absolute score metrics to examine evaluation differences between NLG tasks.
Here, we organize our results into these categories to generalize trends about decoding techniques.
The ratio-based metrics are relative score metrics, since evaluation relies on comparing ratios between demographics.
The latter three metrics are absolute score metrics that have target values of zero indicating no bias.

For the relative score metrics, search and sampling techniques generate similar outcomes.
An interesting result between sampling techniques for the regard metric is that nucleus sampling is less biased yet more negative than top-$k$ sampling.
For the absolute score metrics, we find that beam search is the most unbiased technique, closely followed by greedy search and then top-$k$ and nucleus sampling.
Through our study, we discover that text diversity is not accounted for in any of the bias metrics, yet diversity can be a confounding factor.
Specifically, beam search is the least diverse,\footnote{We report average generated text length and vocabulary sizes to estimate diversity in Appendix Table~\ref{tab:length-vocab}.} followed by greedy search, top-$k$ sampling, then nucleus sampling.
Results indicate that the less diverse search techniques lead to better scores for individual fairness, group fairness, and gendered word co-occurrence ratios.

We hope these experimental results will encourage researchers to document sampling techniques, consider how metrics can be formulated to evaluate both bias and other factors of generation quality, and inspire more comprehensive studies.\footnote{Results are summarized in Appendix Tables \ref{tab:decoding}, \ref{tab:extra-regard}, and \ref{tab:decoding-samples}.}

\subsection{Biases from Evaluation}
Biases can arise from both general evaluations and bias evaluations for NLG tasks.

\paragraph{General Evaluations}
Current standards for NLG evaluation can reinforce certain types of language and penalize others.
For example, using perplexity as measured by models pre-trained on datasets largely containing non-AAE text leads to an unfair evaluation of AAE text. Additionally, the subjectivity of generation tasks means that much of NLG evaluation depends on human labels.
Since humans from different backgrounds are accustomed to different societal norms and linguistic variations, the choice of human annotators could drastically influence the evaluation standards for generated text.

\paragraph{Bias Evaluations}
It is difficult to evaluate societal biases in NLG tasks because NLG can be open-domain, and there are many different notions of biases from various backgrounds and cultures \citep{sambasivan2021fairness}.
These factors lead to the use of a variety of metrics to evaluate biases (Section~\ref{sec:biases-harms}).
To avoid experimental bias in evaluation, we recommend using multiple metrics to cover many types of biases at various granularities.
We identify three points to emphasize the need for more comprehensive evaluations.
First, most existing works on biases in generation center around one demographic dimension (often gender and from a Western perspective, e.g., using standard Western occupations).
While there has been no comprehensive study on whether mitigating biases for one demographic dimension (e.g., gender) may exacerbate biases for others (e.g., race, intersectional identities), this is a possibility we must consider.
Second, most works only evaluate bias through a single intermediate proxy; however, different metrics are defined at different granularities (e.g., sentiment is sentence-level, gendered word ratio is word-level).
Finally, different evaluation datasets test for specific types of biases and are influenced by the backgrounds of the curators.
Collectively evaluating biases across demographic dimensions and granularities can thus help reduce experimentally-biased evaluations. 

\subsection{Biases from Deploying Systems}
\label{ssec:deploying}
In terms of deploying NLG systems,
there is a feedback loop that benefits some communities and further disadvantages others.
While this feedback loop is not unique to NLG systems, these systems that directly interact with users make good cautionary examples.

First, many deployed language technologies require internet access both to use and contribute feedback, thus favoring the views and languages of those privileged with this access.
For example, anyone can contribute feedback to Google Translate, but if contributions and subsequent improvements are focused on high-resource languages, this further increases the accuracy gap between the high and low resource languages, diminishing opportunities for speakers of the low resource languages, i.e., \emph{representation disparity} \citep{hashimoto2018fairness}.

Second, those who are unable to achieve their goals from using these language technologies (e.g., unsuccessful translation, unhelpful or offensive chat bot) are less likely to continue using the technology. 
This means that there is less feedback and data to improve the technologies, reinforcing the decreased effectiveness for certain populations, i.e., \emph{disparity amplification} \citep{hashimoto2018fairness}. 

One way we might intervene is to follow a more targeted approach for data and feedback collection, e.g., from excluded populations.
However, we acknowledge that this remains a difficult task and that it is also necessary to be aware of ``community goals'' and other factors in order to co-design language technologies without inflicting additional harm on marginalized populations \citep{bird-2020-decolonising}.

\section{Progress, Trends, and Challenges}
\label{sec:progress}
Following the discussion of contributors to biases, we survey trends and challenges for reducing biases in NLG.

\subsection{Data Methods} 
Data-based methods for both bias analysis and mitigation use the general idea of counterfactual data augmentation (CDA) \citep{lu2020gender} to curate sets of counterfactual prompts.
A common method for analysis is using targeted prompts to induce NLG models to reveal biases.
For data-based mitigation, existing works focus on fine-tuning large models or training smaller models with datasets that are balanced with respect to targeted demographics.

\paragraph{Curated Datasets}
Existing datasets to study biases in translation include parallel sentences tagged with speaker or subject gender information \citep{vanmassenhove-etal-2018-getting,habash-etal-2019-automatic} and datasets to study gender biases when translating from neutral references of a person (e.g., \textit{nurse} in English, gender-neutral pronouns) to gendered instances (e.g., \textit{enfermera} or \textit{enfermero} in Spanish, gendered pronouns) \citep{cho-etal-2019-measuring,stanovsky-etal-2019-evaluating,gonen-webster-2020-automatically,kocmi-etal-2020-gender}.
\citet{renduchintala2021investigating} additionally provide a dataset to study translation of neutral references in unambiguous contexts.
Other works present parallel corpora of biased versus unbiased framings and presuppositions \citep{pryzant2020automatically} and AAE versus WAE equivalents \citep{groenwold-etal-2020-investigating}.
\citet{sheng2019woman,huang-etal-2020-reducing,dhamala2021bold} additionally curate sets of prompts that can be used to evaluate biases in autocomplete generation.

\paragraph{Bias Analysis}
Most bias analyses of NLG tasks use prompts to probe for different biases in generated text, e.g., regarding social perception \citep{sheng2019woman}, gender in translation \citep{prates2019assessing}, names \citep{shwartz-etal-2020-grounded}, sentiment distribution \citep{huang-etal-2020-reducing}, dialects  \citep{groenwold-etal-2020-investigating}, dialogue personas \citep{sheng2021revealing}, or other notions of similarity across demographics \citep{yeo-chen-2020-defining,henderson2018ethical}.
\citet{vig2020investigating} also use prompts to investigate gender biases, though they do so in the context of a causal mediation analysis.
Furthermore, \citet{prates2019assessing} and \citet{farkas2020measure} compare pronoun gender biases in translations (induced with prompts) to real-world statistics.

\paragraph{Bias Mitigation}
Methods can broadly be classified into two categories based on the type of data applied.
The first category encompasses methods that fine-tune or train on a balanced dataset to lessen the effects of the model relying on spurious correlations between imbalanced data and task performance.
CDA has been applied to datasets used for continued or fresh training in dialogue generation \citep{dinan-etal-2020-queens,liu-etal-2020-gender} as well as machine translation \citep{saunders-byrne-2020-reducing,costa-jussa-de-jorge-2020-fine,stafanovics-etal-2020-mitigating}.
The second category is methods that attach a short prefix at training time \citep{vanmassenhove-etal-2018-getting,basta-etal-2020-towards,alhafni-etal-2020-gender} or inference time \citep{moryossef-etal-2019-filling}.

\paragraph{Challenges}
The size of state-of-the-art pre-trained models and varying definitions of biases in generation present difficulties for creating standardized datasets that are generally effective across biases and demographics.
Moreover, it remains to be seen whether data-based mitigation is as effective for open-domain NLG tasks as it is for more constrained settings.

\subsection{Training Methods}
In addition to data-based mitigation, training-based mitigation is another popular class of methods to reduce biases in generation.

\paragraph{Bias Mitigation}
Several works that use training-based mitigation techniques rely on regularization \citep{bordia-bowman-2019-identifying,qian-etal-2019-reducing,huang-etal-2020-reducing,liu-etal-2020-gender,saunders-byrne-2020-reducing}.
There are also works that induce control by incorporating a bias control code through conditional training \citep{dinan-etal-2020-queens}, by appending a target value to inputs during training \citep{ma-etal-2020-powertransformer}, by using a normative classifier to produce reward values for backpropagation \citep{peng-etal-2020-reducing}, or through adversarial training \citep{liu-etal-2020-mitigating}.
Other techniques include using debiased word embeddings \citep{escude-font-costa-jussa-2019-equalizing}, identifying and editing out subjective words \citep{pryzant2020automatically}, and using Markov random fields to preserve morpho-syntactic agreement during reinflection \citep{zmigrod-etal-2019-counterfactual}.

\paragraph{Challenges}
The main challenge of bias mitigation through training methods is that it is costly and impractical to re-train models for new biases encountered.
In fact, most of the techniques that rely on training from scratch use smaller architectures (exceptions are from larger institutions).

\subsection{Inference Methods}
While the existing literature on inference time methods for bias mitigation is sparse, decoding-based methods are a promising alternative to data- and training-based methods.
Specifically, these methods are compatible with any pre-trained language model for generation without additional training.
Given recent development of inference-time methods for control that can reduce toxicity (e.g., PPLM \citep{dathathri2019plug}, GeDi \citep{krause2020gedi}, DExperts \citep{liu2021fly}), there is potential for extending these methods to bias mitigation.

\paragraph{Bias Mitigation}
For autocomplete and dialogue generation, \citet{sheng-etal-2020-towards} formulate bias triggers using gradient-based methods of \citet{wallace-etal-2019-universal}.
These triggers are appended to prompts during inference time to control text generation to be more equalized towards different demographics.
For translation, \citet{saunders-byrne-2020-reducing} present a lattice rescoring procedure that creates gender-inflected search spaces to rescore text for more accurate translations, and \citet{saunders2021first} subsequently use this lattice structure to present more gendered options during beam search and rerank translation hypotheses according to gender criteria.
For dialogue generation, \citet{sheng2020nice} introduce a constrained decoding method that uses $n$-gram similarity to guide generation away from ad hominems towards marginalized groups.
For autocomplete generation, \citet{schick2021self} present a self-debiasing scheme that re-weights word probabilities to generate less undesirable words.

\paragraph{Challenges}
Control methods at inference time could potentially steer the model into degenerate spaces, so it is important to also evaluate these methods for coherence, fluency, and task relevance.

\subsection{Evaluation Methods}
\label{ssec:progress-evaluation}
There are two types of evaluations: those that rely on absolute scores and those that rely on relative scores.
Absolute score evaluations use an accumulated score to summarize inequalities between demographics, whereas relative evaluations explicitly report inequalities between all demographics.
While it is possible to convert between relative and absolute scores, distinguishing between how existing works choose to portray evaluations allows us to examine differences between generation tasks.

\paragraph{Absolute Evaluations}
We find that the \emph{transformation} class of generation tasks favors bias evaluation through absolute metrics, which is possible because these tasks involve relatively more constrained forms of generation.
Examples of evaluation objectives through absolute scores include \citet{peng-etal-2020-reducing} reducing non-normative generations, \citet{ma-etal-2020-powertransformer} increasing the accuracy of the change in agency, \citet{zmigrod-etal-2019-counterfactual} increasing the number of correct inflections,
\citet{huang-etal-2020-reducing} reducing individual and group fairness scores, and \citet{sheng2020nice} reducing the amount of ad hominems towards marginalized groups. 
Studies of gender bias in machine translation are well-suited to evaluations using absolute scores: many use BLEU and its variants to evaluate correct gender inflections and translations \citep{moryossef-etal-2019-filling,escude-font-costa-jussa-2019-equalizing,elaraby2018gender,habash-etal-2019-automatic,alhafni-etal-2020-gender} or accuracy on WinoMT \citep{saunders-byrne-2020-reducing, saunders-etal-2020-neural,kocmi-etal-2020-gender,costa-jussa-de-jorge-2020-fine,costa2020gender,basta-etal-2020-towards,choubey2021improving,saunders2021first}. 

\paragraph{Relative Evaluations}
In terms of evaluation through relative scores, examples from existing works are mainly from \emph{continuation} generation tasks.
We infer that the less constrained, open-domain nature of continuation generation tasks makes it more preferable to evaluate mitigation through more flexible comparisons rather than absolute scores.
For autocomplete generation, \citet{sheng2019woman,sheng-etal-2020-towards} and \citet{groenwold-etal-2020-investigating} compare regard or sentiment scores across demographics, \citet{shwartz-etal-2020-grounded} compare names across various intermediate metrics,  \citet{vig2020investigating} measure proportional differences between the amount of bias under a gendered versus ambiguous reading, and \citet{yeo-chen-2020-defining} compare occupations generated for different genders.
Bias studies in dialogue generation use relative scores by comparing sentiment and offensive language discrepancies \citep{henderson2018ethical,liu-etal-2020-gender,liu-etal-2020-mitigating} and the percentage of gendered words \citep{dinan-etal-2020-queens}. 

\paragraph{Challenges}

A trade-off between framing biases as a relative or absolute metric is that relative metrics can be more flexibly aligned to normative concerns like social perception. 
Absolute metrics that look for ratios of gendered words or other indicator words assume that there is a set of words that captures all the differences between demographic groups, regardless of whether these differences are related to normative definitions of harm.
There are also absolute metrics such as those of \citet{huang-etal-2020-reducing} that can incorporate intermediate metrics that are more aligned with normative behavior, though these metrics reduce the notion of biases to a single value, which could erase historical inequalities between groups.

\section{Open Problems and Proposals}
\label{sec:open}
As a fairly nascent area of exploration, the study of biases in language generation still poses many challenges.
Throughout this paper, we discuss challenges associated with different components in a generation pipeline.
With a heightened awareness of the relevant body of work, we conclude with recommendations for open problems.

\paragraph{Bias-Aware Data Curation}
Many works have highlighted the harms and problems when collecting training datasets with limited awareness for potential harms.
Since effective models for NLG tasks are correlated with increasing training data sizes, biases in data collection (e.g., English-centric, drawn from popular Western media) remain a major contributor of biases that manifest in generation. 
Additionally, datasets used to study biases in generation can also be limited (e.g., only for binary gender classes).
For more bias-aware data curation, we suggest diversifying datasets to include more viewpoints from various groups.

\paragraph{Understanding Trade-Offs}
Different methods for analysis, mitigation, and evaluation have unique trade-offs.
Existing works have been relatively small-scale and limited to a small number of biases for specific tasks.
Some useful questions to consider when developing methods to study generation biases are whether we can generalize methods to a diverse set of biases and a wide range of contexts.
It is also important to consider formulating metrics that would jointly mitigate biases and preserve other desired text qualities (e.g., diversity, fluency).

\paragraph{Interactive and Continuous Learning}
The difficulties of measuring and mitigating biases in generation can be reduced with a general framework for interactive and continuous learning.
Over time, such a system could learn from diverse opinions of what constitutes ``fair'' versus ``unfair'' generations across tasks.
A unified framework would centralize and highlight the importance of studying biases in generation, as well as fuel the development of a more comprehensive set of evaluations that may be useful for large-scale studies of impact.

\paragraph{Focusing on Negative Impacts}
Section~\ref{sec:biases-harms} discusses how there are very few existing works on biases that explicitly and meaningfully engage with resulting negative impacts, even though these impacts are what motivate reducing biases.
By re-framing efforts on reducing negative impacts rather than biases, we may be able to define metrics and progress that better correlate with reducing harm.
For example, relative framings of bias metrics could better enable metrics to be more aligned with reducing harms for particularly impacted groups.

\section*{Acknowledgments}
We would like to thank Seraphina Goldfarb-Tarrant, Sunipa Dev, Jason Teoh, members of the Plus Lab, and our anonymous reviewers for the many helpful suggestions that went into this paper.

\section*{Ethics and Broader Implications}
In this work, we present a survey and commentary on the progress and challenges for studying societal biases in language generation.

\paragraph{Data}
We do not check the quality of the datasets used to train popular language generation models (due to limited availability and size), though we do briefly mention problems that other works have found regarding using large datasets that have been minimally filtered.
Some of the surveyed datasets and metrics that are used for evaluating biases approximate binary genders using names typical of specific genders, and may be better re-formulated to avoid harms and curate a more accurate representation of different genders.
On the subject of genders, the majority of bias evaluation data also only evaluate for binary genders---we point out this issue in our survey as well.

\paragraph{Techniques}
Most of the techniques surveyed in this work are trained with or bias-tested with data drawn from Western sources or culture, since that is largely the focus of the existing body of work.
We also refer to studies that point out how techniques for bias do not always transfer across cultures.
Our decoding experiments could potentially fuel misuse by giving those with adversarial interests a better understanding of how decoding algorithms could thwart bias metrics, though we believe transparency around these results outweigh the potential for misuse.

\bibliographystyle{acl_natbib}
\bibliography{anthology,acl2021}

\begin{thebibliography}{101}
\expandafter\ifx\csname natexlab\endcsname\relax\def\natexlab#1{#1}\fi

\bibitem[{Abid et~al.(2021)Abid, Farooqi, and Zou}]{abid2021persistent}
Abubakar Abid, Maheen Farooqi, and James Zou. 2021.
\newblock Persistent anti-muslim bias in large language models.
\newblock \emph{arXiv preprint arXiv:2101.05783}.

\bibitem[{Alhafni et~al.(2020)Alhafni, Habash, and
  Bouamor}]{alhafni-etal-2020-gender}
Bashar Alhafni, Nizar Habash, and Houda Bouamor. 2020.
\newblock \href {https://www.aclweb.org/anthology/2020.gebnlp-1.12}
  {Gender-aware reinflection using linguistically enhanced neural models}.
\newblock In \emph{Proceedings of the Second Workshop on Gender Bias in Natural
  Language Processing}, pages 139--150, Barcelona, Spain (Online). Association
  for Computational Linguistics.

\bibitem[{Barocas et~al.(2017)Barocas, Crawford, Shapiro, and
  Wallach}]{barocas2017problem}
Solon Barocas, Kate Crawford, Aaron Shapiro, and Hanna Wallach. 2017.
\newblock The problem with bias: Allocative versus representational harms in
  machine learning.
\newblock In \emph{9th Annual Conference of the Special Interest Group for
  Computing, Information and Society}.

\bibitem[{Basta et~al.(2020)Basta, Costa-juss{\`a}, and
  Fonollosa}]{basta-etal-2020-towards}
Christine Basta, Marta~R. Costa-juss{\`a}, and Jos{\'e} A.~R. Fonollosa. 2020.
\newblock \href {https://doi.org/10.18653/v1/2020.winlp-1.25} {Towards
  mitigating gender bias in a decoder-based neural machine translation model by
  adding contextual information}.
\newblock In \emph{Proceedings of the The Fourth Widening Natural Language
  Processing Workshop}, pages 99--102, Seattle, USA. Association for
  Computational Linguistics.

\bibitem[{Bender and Friedman(2018)}]{bender-friedman-2018-data}
Emily~M. Bender and Batya Friedman. 2018.
\newblock \href {https://doi.org/10.1162/tacl_a_00041} {Data statements for
  natural language processing: Toward mitigating system bias and enabling
  better science}.
\newblock \emph{Transactions of the Association for Computational Linguistics},
  6:587--604.

\bibitem[{Bender et~al.(2021)Bender, Gebru, McMillan-Major, and
  Shmitchell}]{bender2021dangers}
Emily~M Bender, Timnit Gebru, Angelina McMillan-Major, and Shmargaret
  Shmitchell. 2021.
\newblock On the dangers of stochastic parrots: Can language models be too big.
\newblock \emph{Proceedings of FAccT}.

\bibitem[{Bird(2020)}]{bird-2020-decolonising}
Steven Bird. 2020.
\newblock \href {https://www.aclweb.org/anthology/2020.coling-main.313}
  {Decolonising speech and language technology}.
\newblock In \emph{Proceedings of the 28th International Conference on
  Computational Linguistics}, pages 3504--3519, Barcelona, Spain (Online).
  International Committee on Computational Linguistics.

\bibitem[{Blodgett et~al.(2020)Blodgett, Barocas, Daum{\'e}~III, and
  Wallach}]{blodgett-etal-2020-language}
Su~Lin Blodgett, Solon Barocas, Hal Daum{\'e}~III, and Hanna Wallach. 2020.
\newblock \href {https://doi.org/10.18653/v1/2020.acl-main.485} {Language
  (technology) is power: A critical survey of {``}bias{''} in {NLP}}.
\newblock In \emph{Proceedings of the 58th Annual Meeting of the Association
  for Computational Linguistics}, pages 5454--5476, Online. Association for
  Computational Linguistics.

\bibitem[{Bordia and Bowman(2019)}]{bordia-bowman-2019-identifying}
Shikha Bordia and Samuel~R. Bowman. 2019.
\newblock \href {https://doi.org/10.18653/v1/N19-3002} {Identifying and
  reducing gender bias in word-level language models}.
\newblock In \emph{Proceedings of the 2019 Conference of the North {A}merican
  Chapter of the Association for Computational Linguistics: Student Research
  Workshop}, pages 7--15, Minneapolis, Minnesota. Association for Computational
  Linguistics.

\bibitem[{Brown et~al.(2020)Brown, Mann, Ryder, Subbiah, Kaplan, Dhariwal,
  Neelakantan, Shyam, Sastry, Askell et~al.}]{brown2020language}
Tom~B Brown, Benjamin Mann, Nick Ryder, Melanie Subbiah, Jared Kaplan, Prafulla
  Dhariwal, Arvind Neelakantan, Pranav Shyam, Girish Sastry, Amanda Askell,
  et~al. 2020.
\newblock Language models are few-shot learners.
\newblock \emph{arXiv preprint arXiv:2005.14165}.

\bibitem[{Carlini et~al.(2020)Carlini, Tramer, Wallace, Jagielski,
  Herbert-Voss, Lee, Roberts, Brown, Song, Erlingsson
  et~al.}]{carlini2020extracting}
Nicholas Carlini, Florian Tramer, Eric Wallace, Matthew Jagielski, Ariel
  Herbert-Voss, Katherine Lee, Adam Roberts, Tom Brown, Dawn Song, Ulfar
  Erlingsson, et~al. 2020.
\newblock Extracting training data from large language models.
\newblock \emph{arXiv preprint arXiv:2012.07805}.

\bibitem[{Celis and Keswani(2020)}]{celis2020dialect}
L~Elisa Celis and Vijay Keswani. 2020.
\newblock Dialect diversity in text summarization on twitter.
\newblock \emph{arXiv preprint arXiv:2007.07860}.

\bibitem[{Cercas~Curry et~al.(2020)Cercas~Curry, Robertson, and
  Rieser}]{cercas-curry-etal-2020-conversational}
Amanda Cercas~Curry, Judy Robertson, and Verena Rieser. 2020.
\newblock \href {https://www.aclweb.org/anthology/2020.gebnlp-1.7}
  {Conversational assistants and gender stereotypes: Public perceptions and
  desiderata for voice personas}.
\newblock In \emph{Proceedings of the Second Workshop on Gender Bias in Natural
  Language Processing}, pages 72--78, Barcelona, Spain (Online). Association
  for Computational Linguistics.

\bibitem[{Chen et~al.(2020)Chen, Gan, Cheng, Liu, and
  Liu}]{chen-etal-2020-distilling}
Yen-Chun Chen, Zhe Gan, Yu~Cheng, Jingzhou Liu, and Jingjing Liu. 2020.
\newblock \href {https://doi.org/10.18653/v1/2020.acl-main.705} {Distilling
  knowledge learned in {BERT} for text generation}.
\newblock In \emph{Proceedings of the 58th Annual Meeting of the Association
  for Computational Linguistics}, pages 7893--7905, Online. Association for
  Computational Linguistics.

\bibitem[{Cho et~al.(2019)Cho, Kim, Kim, and Kim}]{cho-etal-2019-measuring}
Won~Ik Cho, Ji~Won Kim, Seok~Min Kim, and Nam~Soo Kim. 2019.
\newblock \href {https://doi.org/10.18653/v1/W19-3824} {On measuring gender
  bias in translation of gender-neutral pronouns}.
\newblock In \emph{Proceedings of the First Workshop on Gender Bias in Natural
  Language Processing}, pages 173--181, Florence, Italy. Association for
  Computational Linguistics.

\bibitem[{Cho et~al.(2021)Cho, Kim, Yang, and Kim}]{cho2021towards}
Won~Ik Cho, Jiwon Kim, Jaeyeong Yang, and Nam~Soo Kim. 2021.
\newblock Towards cross-lingual generalization of translation gender bias.
\newblock In \emph{Proceedings of the 2021 ACM Conference on Fairness,
  Accountability, and Transparency}, pages 449--457.

\bibitem[{Choubey et~al.(2021)Choubey, Currey, Mathur, and
  Dinu}]{choubey2021improving}
Prafulla~Kumar Choubey, Anna Currey, Prashant Mathur, and Georgiana Dinu. 2021.
\newblock Improving gender translation accuracy with filtered self-training.
\newblock \emph{arXiv preprint arXiv:2104.07695}.

\bibitem[{Costa-juss{\`a} et~al.(2020)Costa-juss{\`a}, Escolano, Basta,
  Ferrando, Batlle, and Kharitonova}]{costa2020gender}
Marta~R Costa-juss{\`a}, Carlos Escolano, Christine Basta, Javier Ferrando,
  Roser Batlle, and Ksenia Kharitonova. 2020.
\newblock Gender bias in multilingual neural machine translation: The
  architecture matters.
\newblock \emph{arXiv preprint arXiv:2012.13176}.

\bibitem[{Costa-juss{\`a} and de~Jorge(2020)}]{costa-jussa-de-jorge-2020-fine}
Marta~R. Costa-juss{\`a} and Adri{\`a} de~Jorge. 2020.
\newblock \href {https://www.aclweb.org/anthology/2020.gebnlp-1.3} {Fine-tuning
  neural machine translation on gender-balanced datasets}.
\newblock In \emph{Proceedings of the Second Workshop on Gender Bias in Natural
  Language Processing}, pages 26--34, Barcelona, Spain (Online). Association
  for Computational Linguistics.

\bibitem[{Crawford(2017)}]{crawford2017bias}
Kate Crawford. 2017.
\newblock The trouble with bias.
\newblock Keynote at NeurIPS.

\bibitem[{Dai et~al.(2019)Dai, Yang, Yang, Carbonell, Le, and
  Salakhutdinov}]{dai-etal-2019-transformer}
Zihang Dai, Zhilin Yang, Yiming Yang, Jaime Carbonell, Quoc Le, and Ruslan
  Salakhutdinov. 2019.
\newblock \href {https://doi.org/10.18653/v1/P19-1285} {Transformer-{XL}:
  Attentive language models beyond a fixed-length context}.
\newblock In \emph{Proceedings of the 57th Annual Meeting of the Association
  for Computational Linguistics}, pages 2978--2988, Florence, Italy.
  Association for Computational Linguistics.

\bibitem[{Dathathri et~al.(2019)Dathathri, Madotto, Lan, Hung, Frank, Molino,
  Yosinski, and Liu}]{dathathri2019plug}
Sumanth Dathathri, Andrea Madotto, Janice Lan, Jane Hung, Eric Frank, Piero
  Molino, Jason Yosinski, and Rosanne Liu. 2019.
\newblock Plug and play language models: A simple approach to controlled text
  generation.
\newblock In \emph{International Conference on Learning Representations}.

\bibitem[{Devlin et~al.(2019)Devlin, Chang, Lee, and
  Toutanova}]{devlin-etal-2019-bert}
Jacob Devlin, Ming-Wei Chang, Kenton Lee, and Kristina Toutanova. 2019.
\newblock \href {https://doi.org/10.18653/v1/N19-1423} {{BERT}: Pre-training of
  deep bidirectional transformers for language understanding}.
\newblock In \emph{Proceedings of the 2019 Conference of the North {A}merican
  Chapter of the Association for Computational Linguistics: Human Language
  Technologies, Volume 1 (Long and Short Papers)}, pages 4171--4186,
  Minneapolis, Minnesota. Association for Computational Linguistics.

\bibitem[{Dhamala et~al.(2021)Dhamala, Sun, Kumar, Krishna, Pruksachatkun,
  Chang, and Gupta}]{dhamala2021bold}
Jwala Dhamala, Tony Sun, Varun Kumar, Satyapriya Krishna, Yada Pruksachatkun,
  Kai-Wei Chang, and Rahul Gupta. 2021.
\newblock Bold: Dataset and metrics for measuring biases in open-ended language
  generation.
\newblock \emph{Proceedings of FAccT}.

\bibitem[{Dinan et~al.(2020{\natexlab{a}})Dinan, Fan, Williams, Urbanek, Kiela,
  and Weston}]{dinan-etal-2020-queens}
Emily Dinan, Angela Fan, Adina Williams, Jack Urbanek, Douwe Kiela, and Jason
  Weston. 2020{\natexlab{a}}.
\newblock \href {https://doi.org/10.18653/v1/2020.emnlp-main.656} {Queens are
  powerful too: Mitigating gender bias in dialogue generation}.
\newblock In \emph{Proceedings of the 2020 Conference on Empirical Methods in
  Natural Language Processing (EMNLP)}, pages 8173--8188, Online. Association
  for Computational Linguistics.

\bibitem[{Dinan et~al.(2020{\natexlab{b}})Dinan, Fan, Wu, Weston, Kiela, and
  Williams}]{dinan-etal-2020-multi}
Emily Dinan, Angela Fan, Ledell Wu, Jason Weston, Douwe Kiela, and Adina
  Williams. 2020{\natexlab{b}}.
\newblock \href {https://doi.org/10.18653/v1/2020.emnlp-main.23}
  {Multi-dimensional gender bias classification}.
\newblock In \emph{Proceedings of the 2020 Conference on Empirical Methods in
  Natural Language Processing (EMNLP)}, pages 314--331, Online. Association for
  Computational Linguistics.

\bibitem[{Dwork et~al.(2012)Dwork, Hardt, Pitassi, Reingold, and
  Zemel}]{dwork2012fairness}
Cynthia Dwork, Moritz Hardt, Toniann Pitassi, Omer Reingold, and Richard Zemel.
  2012.
\newblock Fairness through awareness.
\newblock In \emph{Proceedings of the 3rd innovations in theoretical computer
  science conference}, pages 214--226.

\bibitem[{Elaraby et~al.(2018)Elaraby, Tawfik, Khaled, Hassan, and
  Osama}]{elaraby2018gender}
Mostafa Elaraby, Ahmed~Y Tawfik, Mahmoud Khaled, Hany Hassan, and Aly Osama.
  2018.
\newblock Gender aware spoken language translation applied to english-arabic.
\newblock In \emph{2018 2nd International Conference on Natural Language and
  Speech Processing (ICNLSP)}, pages 1--6. IEEE.

\bibitem[{Escud{\'e}~Font and
  Costa-juss{\`a}(2019)}]{escude-font-costa-jussa-2019-equalizing}
Joel Escud{\'e}~Font and Marta~R. Costa-juss{\`a}. 2019.
\newblock \href {https://doi.org/10.18653/v1/W19-3821} {Equalizing gender bias
  in neural machine translation with word embeddings techniques}.
\newblock In \emph{Proceedings of the First Workshop on Gender Bias in Natural
  Language Processing}, pages 147--154, Florence, Italy. Association for
  Computational Linguistics.

\bibitem[{Fan et~al.(2018)Fan, Lewis, and Dauphin}]{fan2018hierarchical}
Angela Fan, Mike Lewis, and Yann Dauphin. 2018.
\newblock Hierarchical neural story generation.
\newblock In \emph{Proceedings of the 56th Annual Meeting of the Association
  for Computational Linguistics (Volume 1: Long Papers)}, pages 889--898.

\bibitem[{Farkas and N{\'e}meth(2020)}]{farkas2020measure}
Anna Farkas and Ren{\'a}ta N{\'e}meth. 2020.
\newblock How to measure gender bias in machine translation: Optimal
  translators, multiple reference points.
\newblock \emph{arXiv preprint arXiv:2011.06445}.

\bibitem[{Ferrer et~al.(2021)Ferrer, van Nuenen, Such, and
  Criado}]{ferrer2020discovering}
Xavier Ferrer, Tom van Nuenen, Jose~M Such, and Natalia Criado. 2021.
\newblock Discovering and categorising language biases in reddit.
\newblock In \emph{Proceedings of the International AAAI Conference on Web and
  Social Media}, volume~15.

\bibitem[{Gebru et~al.(2018)Gebru, Morgenstern, Vecchione, Vaughan, Wallach,
  Daum{\'e}~III, and Crawford}]{gebru2018datasheets}
Timnit Gebru, Jamie Morgenstern, Briana Vecchione, Jennifer~Wortman Vaughan,
  Hanna Wallach, Hal Daum{\'e}~III, and Kate Crawford. 2018.
\newblock Datasheets for datasets.
\newblock \emph{arXiv preprint arXiv:1803.09010}.

\bibitem[{Gonen and Webster(2020)}]{gonen-webster-2020-automatically}
Hila Gonen and Kellie Webster. 2020.
\newblock \href {https://doi.org/10.18653/v1/2020.findings-emnlp.180}
  {Automatically identifying gender issues in machine translation using
  perturbations}.
\newblock In \emph{Findings of the Association for Computational Linguistics:
  EMNLP 2020}, pages 1991--1995, Online. Association for Computational
  Linguistics.

\bibitem[{Groenwold et~al.(2020)Groenwold, Ou, Parekh, Honnavalli, Levy, Mirza,
  and Wang}]{groenwold-etal-2020-investigating}
Sophie Groenwold, Lily Ou, Aesha Parekh, Samhita Honnavalli, Sharon Levy, Diba
  Mirza, and William~Yang Wang. 2020.
\newblock \href {https://doi.org/10.18653/v1/2020.emnlp-main.473}
  {Investigating {A}frican-{A}merican {V}ernacular {E}nglish in
  transformer-based text generation}.
\newblock In \emph{Proceedings of the 2020 Conference on Empirical Methods in
  Natural Language Processing (EMNLP)}, pages 5877--5883, Online. Association
  for Computational Linguistics.

\bibitem[{Habash et~al.(2019)Habash, Bouamor, and
  Chung}]{habash-etal-2019-automatic}
Nizar Habash, Houda Bouamor, and Christine Chung. 2019.
\newblock \href {https://doi.org/10.18653/v1/W19-3822} {Automatic gender
  identification and reinflection in {A}rabic}.
\newblock In \emph{Proceedings of the First Workshop on Gender Bias in Natural
  Language Processing}, pages 155--165, Florence, Italy. Association for
  Computational Linguistics.

\bibitem[{Hardt et~al.(2016)Hardt, Price, and Srebro}]{hardt2016equality}
Moritz Hardt, Eric Price, and Nati Srebro. 2016.
\newblock Equality of opportunity in supervised learning.
\newblock In \emph{Advances in neural information processing systems}, pages
  3315--3323.

\bibitem[{Hashimoto et~al.(2018)Hashimoto, Srivastava, Namkoong, and
  Liang}]{hashimoto2018fairness}
Tatsunori Hashimoto, Megha Srivastava, Hongseok Namkoong, and Percy Liang.
  2018.
\newblock Fairness without demographics in repeated loss minimization.
\newblock In \emph{International Conference on Machine Learning}, pages
  1929--1938. PMLR.

\bibitem[{Henderson et~al.(2018)Henderson, Sinha, Angelard-Gontier, Ke, Fried,
  Lowe, and Pineau}]{henderson2018ethical}
Peter Henderson, Koustuv Sinha, Nicolas Angelard-Gontier, Nan~Rosemary Ke,
  Genevieve Fried, Ryan Lowe, and Joelle Pineau. 2018.
\newblock Ethical challenges in data-driven dialogue systems.
\newblock In \emph{Proceedings of the 2018 AAAI/ACM Conference on AI, Ethics,
  and Society}, pages 123--129.

\bibitem[{Holtzman et~al.(2019)Holtzman, Buys, Du, Forbes, and
  Choi}]{holtzman2019curious}
Ari Holtzman, Jan Buys, Li~Du, Maxwell Forbes, and Yejin Choi. 2019.
\newblock The curious case of neural text degeneration.
\newblock In \emph{International Conference on Learning Representations}.

\bibitem[{Hovy et~al.(2020)Hovy, Bianchi, and
  Fornaciari}]{hovy-etal-2020-sound}
Dirk Hovy, Federico Bianchi, and Tommaso Fornaciari. 2020.
\newblock \href {https://doi.org/10.18653/v1/2020.acl-main.154} {{``}you sound
  just like your father{''} commercial machine translation systems include
  stylistic biases}.
\newblock In \emph{Proceedings of the 58th Annual Meeting of the Association
  for Computational Linguistics}, pages 1686--1690, Online. Association for
  Computational Linguistics.

\bibitem[{Huang et~al.(2020)Huang, Zhang, Jiang, Stanforth, Welbl, Rae, Maini,
  Yogatama, and Kohli}]{huang-etal-2020-reducing}
Po-Sen Huang, Huan Zhang, Ray Jiang, Robert Stanforth, Johannes Welbl, Jack
  Rae, Vishal Maini, Dani Yogatama, and Pushmeet Kohli. 2020.
\newblock \href {https://doi.org/10.18653/v1/2020.findings-emnlp.7} {Reducing
  sentiment bias in language models via counterfactual evaluation}.
\newblock In \emph{Findings of the Association for Computational Linguistics:
  EMNLP 2020}, pages 65--83, Online. Association for Computational Linguistics.

\bibitem[{Hutto and Gilbert(2014)}]{hutto2014vader}
Clayton Hutto and Eric Gilbert. 2014.
\newblock Vader: A parsimonious rule-based model for sentiment analysis of
  social media text.
\newblock In \emph{Proceedings of the International AAAI Conference on Web and
  Social Media}, volume~8.

\bibitem[{Kay et~al.(2015)Kay, Matuszek, and Munson}]{kay2015unequal}
Matthew Kay, Cynthia Matuszek, and Sean~A Munson. 2015.
\newblock Unequal representation and gender stereotypes in image search results
  for occupations.
\newblock In \emph{Proceedings of the 33rd Annual ACM Conference on Human
  Factors in Computing Systems}, pages 3819--3828.

\bibitem[{Kiritchenko and Mohammad(2018)}]{kiritchenko2018examining}
Svetlana Kiritchenko and Saif Mohammad. 2018.
\newblock Examining gender and race bias in two hundred sentiment analysis
  systems.
\newblock In \emph{Proceedings of the Seventh Joint Conference on Lexical and
  Computational Semantics}, pages 43--53.

\bibitem[{Kiritchenko et~al.(2020)Kiritchenko, Nejadgholi, and
  Fraser}]{kiritchenko2020confronting}
Svetlana Kiritchenko, Isar Nejadgholi, and Kathleen~C Fraser. 2020.
\newblock Confronting abusive language online: A survey from the ethical and
  human rights perspective.
\newblock \emph{arXiv preprint arXiv:2012.12305}.

\bibitem[{Kirk et~al.(2021)Kirk, Jun, Iqbal, Benussi, Volpin, Dreyer,
  Shtedritski, and Asano}]{kirk2021true}
Hannah Kirk, Yennie Jun, Haider Iqbal, Elias Benussi, Filippo Volpin,
  Frederic~A Dreyer, Aleksandar Shtedritski, and Yuki~M Asano. 2021.
\newblock How true is gpt-2? an empirical analysis of intersectional
  occupational biases.
\newblock \emph{arXiv preprint arXiv:2102.04130}.

\bibitem[{Kocmi et~al.(2020)Kocmi, Limisiewicz, and
  Stanovsky}]{kocmi-etal-2020-gender}
Tom Kocmi, Tomasz Limisiewicz, and Gabriel Stanovsky. 2020.
\newblock \href {https://www.aclweb.org/anthology/2020.wmt-1.39} {Gender
  coreference and bias evaluation at {WMT} 2020}.
\newblock In \emph{Proceedings of the Fifth Conference on Machine Translation},
  pages 357--364, Online. Association for Computational Linguistics.

\bibitem[{Krause et~al.(2020)Krause, Gotmare, McCann, Keskar, Joty, Socher, and
  Rajani}]{krause2020gedi}
Ben Krause, Akhilesh~Deepak Gotmare, Bryan McCann, Nitish~Shirish Keskar,
  Shafiq Joty, Richard Socher, and Nazneen~Fatema Rajani. 2020.
\newblock Gedi: Generative discriminator guided sequence generation.
\newblock \emph{arXiv preprint arXiv:2009.06367}.

\bibitem[{Levy et~al.(2021)Levy, Saxon, and Wang}]{levy2021truth}
Sharon Levy, Michael Saxon, and William~Yang Wang. 2021.
\newblock The truth is out there: Investigating conspiracy theories in text
  generation.
\newblock In \emph{Findings of The Joint Conference of the 59th Annual Meeting
  of the Association for Computational Linguistics and the 11th International
  Joint Conference on Natural Language Processing}.

\bibitem[{Liu et~al.(2021)Liu, Sap, Lu, Swayamdipta, Bhagavatula, Smith, and
  Choi}]{liu2021fly}
Alisa Liu, Maarten Sap, Ximing Lu, Swabha Swayamdipta, Chandra Bhagavatula,
  Noah~A. Smith, and Yejin Choi. 2021.
\newblock \href {https://arxiv.org/abs/2105.03023} {{DExperts}: Decoding-time
  controlled text generation with experts and anti-experts}.
\newblock In \emph{Proceedings of the 59th Annual Meeting of the Association
  for Computational Linguistics}.

\bibitem[{Liu et~al.(2020{\natexlab{a}})Liu, Dacon, Fan, Liu, Liu, and
  Tang}]{liu-etal-2020-gender}
Haochen Liu, Jamell Dacon, Wenqi Fan, Hui Liu, Zitao Liu, and Jiliang Tang.
  2020{\natexlab{a}}.
\newblock \href {https://www.aclweb.org/anthology/2020.coling-main.390} {Does
  gender matter? towards fairness in dialogue systems}.
\newblock In \emph{Proceedings of the 28th International Conference on
  Computational Linguistics}, pages 4403--4416, Barcelona, Spain (Online).
  International Committee on Computational Linguistics.

\bibitem[{Liu et~al.(2020{\natexlab{b}})Liu, Wang, Wang, Liu, Liu, and
  Tang}]{liu-etal-2020-mitigating}
Haochen Liu, Wentao Wang, Yiqi Wang, Hui Liu, Zitao Liu, and Jiliang Tang.
  2020{\natexlab{b}}.
\newblock \href {https://doi.org/10.18653/v1/2020.emnlp-main.64} {Mitigating
  gender bias for neural dialogue generation with adversarial learning}.
\newblock In \emph{Proceedings of the 2020 Conference on Empirical Methods in
  Natural Language Processing (EMNLP)}, pages 893--903, Online. Association for
  Computational Linguistics.

\bibitem[{Lu et~al.(2020)Lu, Mardziel, Wu, Amancharla, and
  Datta}]{lu2020gender}
Kaiji Lu, Piotr Mardziel, Fangjing Wu, Preetam Amancharla, and Anupam Datta.
  2020.
\newblock Gender bias in neural natural language processing.
\newblock In \emph{Logic, Language, and Security}, pages 189--202. Springer.

\bibitem[{Lucy and Bamman(2021)}]{lucy2021gender}
Li~Lucy and David Bamman. 2021.
\newblock Gender and representation bias in gpt-3 generated stories.
\newblock In \emph{Proceedings of the Third Workshop on Narrative
  Understanding}, pages 48--55.

\bibitem[{Ma et~al.(2020)Ma, Sap, Rashkin, and
  Choi}]{ma-etal-2020-powertransformer}
Xinyao Ma, Maarten Sap, Hannah Rashkin, and Yejin Choi. 2020.
\newblock \href {https://doi.org/10.18653/v1/2020.emnlp-main.602}
  {{P}ower{T}ransformer: Unsupervised controllable revision for biased language
  correction}.
\newblock In \emph{Proceedings of the 2020 Conference on Empirical Methods in
  Natural Language Processing (EMNLP)}, pages 7426--7441, Online. Association
  for Computational Linguistics.

\bibitem[{Moryossef et~al.(2019)Moryossef, Aharoni, and
  Goldberg}]{moryossef-etal-2019-filling}
Amit Moryossef, Roee Aharoni, and Yoav Goldberg. 2019.
\newblock \href {https://doi.org/10.18653/v1/W19-3807} {Filling gender {\&}
  number gaps in neural machine translation with black-box context injection}.
\newblock In \emph{Proceedings of the First Workshop on Gender Bias in Natural
  Language Processing}, pages 49--54, Florence, Italy. Association for
  Computational Linguistics.

\bibitem[{Nozza et~al.(2021)Nozza, Bianchi, and Hovy}]{nozza2021honest}
Debora Nozza, Federico Bianchi, and Dirk Hovy. 2021.
\newblock Honest: Measuring hurtful sentence completion in language models.
\newblock In \emph{Proceedings of the 2021 Conference of the North American
  Chapter of the Association for Computational Linguistics: Human Language
  Technologies}, pages 2398--2406.

\bibitem[{Ong(2017)}]{ong2017facebook}
Thuy Ong. 2017.
\newblock \href
  {https://www.theverge.com/us-world/2017/10/24/16533496/facebook-apology-wrong-translation-palestinian-arrested-post-good-morning}
  {\emph{Facebook apologizes after wrong translation sees Palestinian man
  arrested for posting 'good morning'}}.

\bibitem[{Park et~al.(2018)Park, Shin, and Fung}]{park2018reducing}
Ji~Ho Park, Jamin Shin, and Pascale Fung. 2018.
\newblock Reducing gender bias in abusive language detection.
\newblock In \emph{Proceedings of the 2018 Conference on Empirical Methods in
  Natural Language Processing}, pages 2799--2804.

\bibitem[{Paullada et~al.(2020)Paullada, Raji, Bender, Denton, and
  Hanna}]{paullada2020data}
Amandalynne Paullada, Inioluwa~Deborah Raji, Emily~M Bender, Emily Denton, and
  Alex Hanna. 2020.
\newblock Data and its (dis) contents: A survey of dataset development and use
  in machine learning research.
\newblock \emph{arXiv preprint arXiv:2012.05345}.

\bibitem[{Peng et~al.(2020)Peng, Li, Frazier, and
  Riedl}]{peng-etal-2020-reducing}
Xiangyu Peng, Siyan Li, Spencer Frazier, and Mark Riedl. 2020.
\newblock \href {https://www.aclweb.org/anthology/2020.inlg-1.43} {Reducing
  non-normative text generation from language models}.
\newblock In \emph{Proceedings of the 13th International Conference on Natural
  Language Generation}, pages 374--383, Dublin, Ireland. Association for
  Computational Linguistics.

\bibitem[{Prates et~al.(2019)Prates, Avelar, and Lamb}]{prates2019assessing}
Marcelo~OR Prates, Pedro~H Avelar, and Lu{\'\i}s~C Lamb. 2019.
\newblock Assessing gender bias in machine translation: a case study with
  google translate.
\newblock \emph{Neural Computing and Applications}, pages 1--19.

\bibitem[{Pryzant et~al.(2020)Pryzant, Martinez, Dass, Kurohashi, Jurafsky, and
  Yang}]{pryzant2020automatically}
Reid Pryzant, Richard~Diehl Martinez, Nathan Dass, Sadao Kurohashi, Dan
  Jurafsky, and Diyi Yang. 2020.
\newblock Automatically neutralizing subjective bias in text.
\newblock In \emph{Proceedings of the AAAI Conference on Artificial
  Intelligence}, volume~34, pages 480--489.

\bibitem[{Qian et~al.(2019)Qian, Muaz, Zhang, and
  Hyun}]{qian-etal-2019-reducing}
Yusu Qian, Urwa Muaz, Ben Zhang, and Jae~Won Hyun. 2019.
\newblock \href {https://doi.org/10.18653/v1/P19-2031} {Reducing gender bias in
  word-level language models with a gender-equalizing loss function}.
\newblock In \emph{Proceedings of the 57th Annual Meeting of the Association
  for Computational Linguistics: Student Research Workshop}, pages 223--228,
  Florence, Italy. Association for Computational Linguistics.

\bibitem[{Radford et~al.(2018)Radford, Narasimhan, Salimans, and
  Sutskever}]{radford2018improving}
Alec Radford, Karthik Narasimhan, Tim Salimans, and Ilya Sutskever. 2018.
\newblock Improving language understanding by generative pre-training.

\bibitem[{Radford et~al.(2019)Radford, Wu, Child, Luan, Amodei, and
  Sutskever}]{radford2019language}
Alec Radford, Jeffrey Wu, Rewon Child, David Luan, Dario Amodei, and Ilya
  Sutskever. 2019.
\newblock Language models are unsupervised multitask learners.
\newblock \emph{OpenAI blog}, 1(8):9.

\bibitem[{Raffel et~al.(2020)Raffel, Shazeer, Roberts, Lee, Narang, Matena,
  Zhou, Li, and Liu}]{raffel2020exploring}
Colin Raffel, Noam Shazeer, Adam Roberts, Katherine Lee, Sharan Narang, Michael
  Matena, Yanqi Zhou, Wei Li, and Peter~J Liu. 2020.
\newblock Exploring the limits of transfer learning with a unified text-to-text
  transformer.
\newblock \emph{Journal of Machine Learning Research}, 21:1--67.

\bibitem[{Renduchintala and Williams(2021)}]{renduchintala2021investigating}
Adithya Renduchintala and Adina Williams. 2021.
\newblock Investigating failures of automatic translation in the case of
  unambiguous gender.
\newblock \emph{arXiv preprint arXiv:2104.07838}.

\bibitem[{Roberts et~al.(2020)Roberts, Liang, Neubig, and
  Lipton}]{roberts2020decoding}
Nicholas Roberts, Davis Liang, Graham Neubig, and Zachary~C Lipton. 2020.
\newblock Decoding and diversity in machine translation.
\newblock \emph{arXiv preprint arXiv:2011.13477}.

\bibitem[{Sambasivan et~al.(2021)Sambasivan, Arnesen, Hutchinson, Doshi, and
  Prabhakaran}]{sambasivan2021fairness}
Nithya Sambasivan, Erin Arnesen, Ben Hutchinson, Tulsee Doshi, and Vinodkumar
  Prabhakaran. 2021.
\newblock Re-imagining algorithmic fairness in india and beyond.
\newblock \emph{Proceedings of FAccT}.

\bibitem[{Saunders and Byrne(2020)}]{saunders-byrne-2020-reducing}
Danielle Saunders and Bill Byrne. 2020.
\newblock \href {https://doi.org/10.18653/v1/2020.acl-main.690} {Reducing
  gender bias in neural machine translation as a domain adaptation problem}.
\newblock In \emph{Proceedings of the 58th Annual Meeting of the Association
  for Computational Linguistics}, pages 7724--7736, Online. Association for
  Computational Linguistics.

\bibitem[{Saunders et~al.(2020)Saunders, Sallis, and
  Byrne}]{saunders-etal-2020-neural}
Danielle Saunders, Rosie Sallis, and Bill Byrne. 2020.
\newblock \href {https://www.aclweb.org/anthology/2020.gebnlp-1.4} {Neural
  machine translation doesn{'}t translate gender coreference right unless you
  make it}.
\newblock In \emph{Proceedings of the Second Workshop on Gender Bias in Natural
  Language Processing}, pages 35--43, Barcelona, Spain (Online). Association
  for Computational Linguistics.

\bibitem[{Saunders et~al.(2021)Saunders, Sallis, and Byrne}]{saunders2021first}
Danielle Saunders, Rosie Sallis, and Bill Byrne. 2021.
\newblock First the worst: Finding better gender translations during beam
  search.
\newblock \emph{arXiv preprint arXiv:2104.07429}.

\bibitem[{Savoldi et~al.(2021)Savoldi, Gaido, Bentivogli, Negri, and
  Turchi}]{savoldi2021gender}
Beatrice Savoldi, Marco Gaido, Luisa Bentivogli, Matteo Negri, and Marco
  Turchi. 2021.
\newblock Gender bias in machine translation.
\newblock In \emph{Transactions of the Association for Computational
  Linguistics}.

\bibitem[{Schick et~al.(2021)Schick, Udupa, and Sch{\"u}tze}]{schick2021self}
Timo Schick, Sahana Udupa, and Hinrich Sch{\"u}tze. 2021.
\newblock Self-diagnosis and self-debiasing: A proposal for reducing
  corpus-based bias in nlp.
\newblock \emph{arXiv preprint arXiv:2103.00453}.

\bibitem[{Shah et~al.(2020)Shah, Schwartz, and
  Hovy}]{shah-etal-2020-predictive}
Deven~Santosh Shah, H.~Andrew Schwartz, and Dirk Hovy. 2020.
\newblock \href {https://doi.org/10.18653/v1/2020.acl-main.468} {Predictive
  biases in natural language processing models: A conceptual framework and
  overview}.
\newblock In \emph{Proceedings of the 58th Annual Meeting of the Association
  for Computational Linguistics}, pages 5248--5264, Online. Association for
  Computational Linguistics.

\bibitem[{Sheng et~al.(2021{\natexlab{a}})Sheng, Arnold, Yu, Chang, and
  Peng}]{sheng2021revealing}
Emily Sheng, Josh Arnold, Zhou Yu, Kai-Wei Chang, and Nanyun Peng.
  2021{\natexlab{a}}.
\newblock Revealing persona biases in dialogue systems.
\newblock \emph{arXiv preprint arXiv:2104.08728}.

\bibitem[{Sheng et~al.(2019)Sheng, Chang, Natarajan, and Peng}]{sheng2019woman}
Emily Sheng, Kai-Wei Chang, Prem Natarajan, and Nanyun Peng. 2019.
\newblock The woman worked as a babysitter: On biases in language generation.
\newblock In \emph{Proceedings of the 2019 Conference on Empirical Methods in
  Natural Language Processing and the 9th International Joint Conference on
  Natural Language Processing (EMNLP-IJCNLP)}, pages 3398--3403.

\bibitem[{Sheng et~al.(2020)Sheng, Chang, Natarajan, and
  Peng}]{sheng-etal-2020-towards}
Emily Sheng, Kai-Wei Chang, Prem Natarajan, and Nanyun Peng. 2020.
\newblock \href {https://doi.org/10.18653/v1/2020.findings-emnlp.291} {Towards
  {C}ontrollable {B}iases in {L}anguage {G}eneration}.
\newblock In \emph{Findings of the Association for Computational Linguistics:
  EMNLP 2020}, pages 3239--3254, Online. Association for Computational
  Linguistics.

\bibitem[{Sheng et~al.(2021{\natexlab{b}})Sheng, Chang, Natarajan, and
  Peng}]{sheng2020nice}
Emily Sheng, Kai-Wei Chang, Premkumar Natarajan, and Nanyun Peng.
  2021{\natexlab{b}}.
\newblock "nice try, kiddo": Investigating ad hominems in dialogue responses.
\newblock In \emph{Proceedings of the 2021 Conference of the North American
  Chapter of the Association for Computational Linguistics: Human Language
  Technologies}.

\bibitem[{Sheng and Uthus(2020)}]{sheng-uthus-2020-investigating}
Emily Sheng and David Uthus. 2020.
\newblock \href {https://www.aclweb.org/anthology/2020.gebnlp-1.9}
  {Investigating societal biases in a poetry composition system}.
\newblock In \emph{Proceedings of the Second Workshop on Gender Bias in Natural
  Language Processing}, pages 93--106, Barcelona, Spain (Online). Association
  for Computational Linguistics.

\bibitem[{Shwartz et~al.(2020)Shwartz, Rudinger, and
  Tafjord}]{shwartz-etal-2020-grounded}
Vered Shwartz, Rachel Rudinger, and Oyvind Tafjord. 2020.
\newblock \href {https://doi.org/10.18653/v1/2020.emnlp-main.556} {{``}you are
  grounded!{''}: Latent name artifacts in pre-trained language models}.
\newblock In \emph{Proceedings of the 2020 Conference on Empirical Methods in
  Natural Language Processing (EMNLP)}, pages 6850--6861, Online. Association
  for Computational Linguistics.

\bibitem[{Silva et~al.(2021)Silva, Tambwekar, and Gombolay}]{silva2021towards}
Andrew Silva, Pradyumna Tambwekar, and Matthew Gombolay. 2021.
\newblock Towards a comprehensive understanding and accurate evaluation of
  societal biases in pre-trained transformers.
\newblock In \emph{Proceedings of the 2021 Conference of the North American
  Chapter of the Association for Computational Linguistics: Human Language
  Technologies}, pages 2383--2389.

\bibitem[{Solaiman et~al.(2019)Solaiman, Brundage, Clark, Askell, Herbert-Voss,
  Wu, Radford, Krueger, Kim, Kreps et~al.}]{solaiman2019release}
Irene Solaiman, Miles Brundage, Jack Clark, Amanda Askell, Ariel Herbert-Voss,
  Jeff Wu, Alec Radford, Gretchen Krueger, Jong~Wook Kim, Sarah Kreps, et~al.
  2019.
\newblock Release strategies and the social impacts of language models.
\newblock \emph{arXiv preprint arXiv:1908.09203}.

\bibitem[{Stafanovi{\v{c}}s et~al.(2020)Stafanovi{\v{c}}s, Pinnis, and
  Bergmanis}]{stafanovics-etal-2020-mitigating}
Art{\=u}rs Stafanovi{\v{c}}s, M{\=a}rcis Pinnis, and Toms Bergmanis. 2020.
\newblock \href {https://www.aclweb.org/anthology/2020.wmt-1.73} {Mitigating
  gender bias in machine translation with target gender annotations}.
\newblock In \emph{Proceedings of the Fifth Conference on Machine Translation},
  pages 629--638, Online. Association for Computational Linguistics.

\bibitem[{Stanovsky et~al.(2019)Stanovsky, Smith, and
  Zettlemoyer}]{stanovsky-etal-2019-evaluating}
Gabriel Stanovsky, Noah~A. Smith, and Luke Zettlemoyer. 2019.
\newblock \href {https://doi.org/10.18653/v1/P19-1164} {Evaluating gender bias
  in machine translation}.
\newblock In \emph{Proceedings of the 57th Annual Meeting of the Association
  for Computational Linguistics}, pages 1679--1684, Florence, Italy.
  Association for Computational Linguistics.

\bibitem[{Sun et~al.(2019)Sun, Gaut, Tang, Huang, ElSherief, Zhao, Mirza,
  Belding, Chang, and Wang}]{sun-etal-2019-mitigating}
Tony Sun, Andrew Gaut, Shirlyn Tang, Yuxin Huang, Mai ElSherief, Jieyu Zhao,
  Diba Mirza, Elizabeth Belding, Kai-Wei Chang, and William~Yang Wang. 2019.
\newblock \href {https://doi.org/10.18653/v1/P19-1159} {Mitigating gender bias
  in natural language processing: Literature review}.
\newblock In \emph{Proceedings of the 57th Annual Meeting of the Association
  for Computational Linguistics}, pages 1630--1640, Florence, Italy.
  Association for Computational Linguistics.

\bibitem[{Sun et~al.(2021)Sun, Webster, Shah, Wang, and Johnson}]{sun2021they}
Tony Sun, Kellie Webster, Apu Shah, William~Yang Wang, and Melvin Johnson.
  2021.
\newblock They, them, theirs: Rewriting with gender-neutral english.
\newblock \emph{arXiv preprint arXiv:2102.06788}.

\bibitem[{Tamkin et~al.(2021)Tamkin, Brundage, Clark, and
  Ganguli}]{tamkin2021understanding}
Alex Tamkin, Miles Brundage, Jack Clark, and Deep Ganguli. 2021.
\newblock Understanding the capabilities, limitations, and societal impact of
  large language models.
\newblock \emph{arXiv preprint arXiv:2102.02503}.

\bibitem[{Tomalin et~al.(2021)Tomalin, Byrne, Concannon, Saunders, and
  Ullmann}]{tomalin2021practical}
Marcus Tomalin, Bill Byrne, Shauna Concannon, Danielle Saunders, and Stefanie
  Ullmann. 2021.
\newblock The practical ethics of bias reduction in machine translation: why
  domain adaptation is better than data debiasing.
\newblock \emph{Ethics and Information Technology}, pages 1--15.

\bibitem[{Vanmassenhove et~al.(2018)Vanmassenhove, Hardmeier, and
  Way}]{vanmassenhove-etal-2018-getting}
Eva Vanmassenhove, Christian Hardmeier, and Andy Way. 2018.
\newblock \href {https://doi.org/10.18653/v1/D18-1334} {Getting gender right in
  neural machine translation}.
\newblock In \emph{Proceedings of the 2018 Conference on Empirical Methods in
  Natural Language Processing}, pages 3003--3008, Brussels, Belgium.
  Association for Computational Linguistics.

\bibitem[{Vaswani et~al.(2017)Vaswani, Shazeer, Parmar, Uszkoreit, Jones,
  Gomez, Kaiser, and Polosukhin}]{vaswani2017attention}
Ashish Vaswani, Noam Shazeer, Niki Parmar, Jakob Uszkoreit, Llion Jones,
  Aidan~N Gomez, {\L}ukasz Kaiser, and Illia Polosukhin. 2017.
\newblock Attention is all you need.
\newblock In \emph{Advances in neural information processing systems}, pages
  5998--6008.

\bibitem[{Vig et~al.(2020)Vig, Gehrmann, Belinkov, Qian, Nevo, Singer, and
  Shieber}]{vig2020investigating}
Jesse Vig, Sebastian Gehrmann, Yonatan Belinkov, Sharon Qian, Daniel Nevo,
  Yaron Singer, and Stuart Shieber. 2020.
\newblock Investigating gender bias in language models using causal mediation
  analysis.
\newblock \emph{Advances in Neural Information Processing Systems}, 33.

\bibitem[{Wallace et~al.(2019)Wallace, Feng, Kandpal, Gardner, and
  Singh}]{wallace-etal-2019-universal}
Eric Wallace, Shi Feng, Nikhil Kandpal, Matt Gardner, and Sameer Singh. 2019.
\newblock \href {https://doi.org/10.18653/v1/D19-1221} {Universal adversarial
  triggers for attacking and analyzing {NLP}}.
\newblock In \emph{Proceedings of the 2019 Conference on Empirical Methods in
  Natural Language Processing and the 9th International Joint Conference on
  Natural Language Processing (EMNLP-IJCNLP)}, pages 2153--2162, Hong Kong,
  China. Association for Computational Linguistics.

\bibitem[{Wang and Cho(2019)}]{wang-cho-2019-bert}
Alex Wang and Kyunghyun Cho. 2019.
\newblock \href {https://doi.org/10.18653/v1/W19-2304} {{BERT} has a mouth, and
  it must speak: {BERT} as a {M}arkov random field language model}.
\newblock In \emph{Proceedings of the Workshop on Methods for Optimizing and
  Evaluating Neural Language Generation}, pages 30--36, Minneapolis, Minnesota.
  Association for Computational Linguistics.

\bibitem[{Yang et~al.(2019)Yang, Dai, Yang, Carbonell, Salakhutdinov, and
  Le}]{yang2019xlnet}
Zhilin Yang, Zihang Dai, Yiming Yang, Jaime Carbonell, Russ~R Salakhutdinov,
  and Quoc~V Le. 2019.
\newblock Xlnet: Generalized autoregressive pretraining for language
  understanding.
\newblock In \emph{Advances in neural information processing systems}, pages
  5753--5763.

\bibitem[{Yeo and Chen(2020)}]{yeo-chen-2020-defining}
Catherine Yeo and Alyssa Chen. 2020.
\newblock \href {https://doi.org/10.18653/v1/2020.winlp-1.27} {Defining and
  evaluating fair natural language generation}.
\newblock In \emph{Proceedings of the The Fourth Widening Natural Language
  Processing Workshop}, pages 107--109, Seattle, USA. Association for
  Computational Linguistics.

\bibitem[{Zhang et~al.(2020)Zhang, Sun, Galley, Chen, Brockett, Gao, Gao, Liu,
  and Dolan}]{zhang-etal-2020-dialogpt}
Yizhe Zhang, Siqi Sun, Michel Galley, Yen-Chun Chen, Chris Brockett, Xiang Gao,
  Jianfeng Gao, Jingjing Liu, and Bill Dolan. 2020.
\newblock \href {https://doi.org/10.18653/v1/2020.acl-demos.30} {{DIALOGPT} :
  Large-scale generative pre-training for conversational response generation}.
\newblock In \emph{Proceedings of the 58th Annual Meeting of the Association
  for Computational Linguistics: System Demonstrations}, pages 270--278,
  Online. Association for Computational Linguistics.

\bibitem[{Zhao et~al.(2018)Zhao, Zhou, Li, Wang, and
  Chang}]{zhao-etal-2018-learning}
Jieyu Zhao, Yichao Zhou, Zeyu Li, Wei Wang, and Kai-Wei Chang. 2018.
\newblock \href {https://doi.org/10.18653/v1/D18-1521} {Learning gender-neutral
  word embeddings}.
\newblock In \emph{Proceedings of the 2018 Conference on Empirical Methods in
  Natural Language Processing}, pages 4847--4853, Brussels, Belgium.
  Association for Computational Linguistics.

\bibitem[{Zmigrod et~al.(2019)Zmigrod, Mielke, Wallach, and
  Cotterell}]{zmigrod-etal-2019-counterfactual}
Ran Zmigrod, Sabrina~J. Mielke, Hanna Wallach, and Ryan Cotterell. 2019.
\newblock \href {https://doi.org/10.18653/v1/P19-1161} {Counterfactual data
  augmentation for mitigating gender stereotypes in languages with rich
  morphology}.
\newblock In \emph{Proceedings of the 57th Annual Meeting of the Association
  for Computational Linguistics}, pages 1651--1661, Florence, Italy.
  Association for Computational Linguistics.

\end{thebibliography}

\clearpage
\appendix
\section{Appendices}
\label{sec:appendix}

\subsection{Evaluating Biases Across Decoding Techniques and Metrics}
\label{ssec:decoding-details}

To gain more insight into biases from different decoding techniques, we examine autocomplete generations from GPT (110M params), GPT-2 (small, 117M params), and XLNet (base, 110M params), using the decoding techniques described in Section~\ref{ssec:decoding} through the Transformers\footnote{\url{https://huggingface.co/transformers}} library.
We use standard parameters of $b=16$ for beam search, $k=40$ with a temperature of $0.7$ for top-$k$ sampling, and $p=0.95$ for nucleus sampling \citep{holtzman2019curious}.
In terms of bias metrics, we use existing NLG bias metrics: regard ratio \citep{sheng2019woman}, sentiment ratio \citep{groenwold-etal-2020-investigating}, individual and group fairness through sentiment (IF/GF) \citep{huang-etal-2020-reducing}, and a gendered word co-occurrence scores \citep{bordia-bowman-2019-identifying}.
For all sentiment scores, we use the rule-based sentiment analyzer, VADER \citep{hutto2014vader}.\footnote{\citet{kiritchenko2018examining} show that sentiment classifiers can exhibit biases. We use VADER since 1) it does not rely on learned associations and thus may be less prone to biases, and 2) it has been used to measure biases in previous works \citep{sheng2019woman,groenwold-etal-2020-investigating}.}
We run all our experiments on an RTX 2080Ti GPU.
Generation takes from a couple of minutes to a few hours, depending on the number of samples generated.

\paragraph{Regard Ratios}
\citet{sheng2019woman} introduce 10 prompts to induce occupation- and respect-related generations (e.g., \textit{[BLANK] worked as}, \textit{[BLANK] was thought of as}) and six demographics (\textit{Black}, \textit{White}, \textit{man}, \textit{woman}, \textit{gay}, \textit{straight}) to fill in the \textit{[BLANK]}, for a total of 60 prompts.
The authors define regard as the social perception \textit{towards} a demographic, collect human annotations, and release a BERT-based regard classifier.\footnote{\url{https://github.com/ewsheng/nlg-bias}} 
We follow the original work in reporting percentages of negative, neutral, and positive regard scores per demographic.
For the deterministic search methods, we do not report scores since there are only 10 samples per demographic.
For the stochastic sampling methods, we generate 1000 samples per demographic.
Additionally, we use the regard classifier released by the authors for our evaluations---while we acknowledge that this classifier could also have biases, we believe it is still worthwhile to use it to compare text generated from different decoding techniques.

\paragraph{Sentiment Ratios for AAE and WAE Prompts}
\citet{groenwold-etal-2020-investigating} curate a parallel set of 2,019 AAE and 2,019 WAE prompts and use sentiment classifiers to label text generated from the prompts.
Similar to \citet{sheng2019woman}, this work also reports percentages of negative, neutral, and positive scores.
The VADER sentiment analyzer that we use reports scores in the range of [-1, 1].
When reporting ratios, we use splits recommended by the authors \citep{hutto2014vader} to categorize sentiment values into negative (value$<=$$-0.05$), neutral ($-0.05$$<$value$<$$0.05$), and positive (value$>=$$0.05$) bins.
When reporting average values, we calculate from the unrounded scores from VADER.
We generate one sample per prompt for all decoding techniques.

\paragraph{Individual and Group Fairness Through Sentiment}
\citet{huang-etal-2020-reducing} evaluate fairness across countries, occupations, and genders (binary, as defined through Western names typical of a gender) by first defining 10 templates per dimension (e.g., \textit{People from [BLANK] are}).
For each dimension, they also define a list of dimension instances (e.g., \textit{Syria} as a country) to fill in the \textit{[BLANK]}.
In total, there are 730 prompts across the three attributes.
For our experiments, we generate one sample per prompt.

The authors define the \emph{individual fairness} metric by ``...averaging the Wasserstein-1 distance between the sentiment score distribution of every evaluation sentence and each of its counterfactual sentences across all templates.''
For example, we would compute the distance between the sentiment distributions of the text generated from the template \textit{People from [BLANK] are} for each of the country choices for [\textit{BLANK]}, and sum up the distance scores for all pairs across all templates.

For \emph{group fairness}, the authors  calculate the average of the ``Wasserstein-1 distance between the sentiment distributions of all generated sentences of inputs from [a] subgroup, and that over the entire evaluation set''.
Here, a subgroup means each country, occupation, or binary gender.
For example, we compare the distance between the sentiment distribution of text generated for \textit{Syria} (across all templates) and the sentiment distribution  of text generated for all countries. 

We use \citet{huang-etal-2020-reducing}'s prefix templates and fairness metrics exactly as defined in the original work, so we refer readers to the original work for more details.

\paragraph{Gendered Word Co-occurrence Scores}
This score is based on the one proposed by \citet{bordia-bowman-2019-identifying}, though we use different gendered word lists and evaluate over all text  generated for the other bias metrics, downsampling if necessary so that the amount and sources of generated text are consistent across decoding techniques.
First, we obtain the lists of female words and male words from \citet{zhao-etal-2018-learning} and add gendered pronouns (\textit{he}, \textit{she}, \textit{his}, \textit{him}, \textit{her}) to the respective lists.
For each word in the aggregated sample set, we calculate the probability of the word given any of the female words (in a context window of 20 words before and after a word) and similarly the probability of the word given any of the male words.
We then take the absolute value of the log ratio of the first probability to the second, and report the average and standard deviation across all non-gendered words.
More concretely, given the set of female gendered words $f$, the set of male gendered words $m$, unique non-gendered words $w \in \mathcal{W}$ in a dataset, and the probability of a word given any of the set $g$ of gendered words $\mathcal{P}(w|g)$, we calculate the mean
\begin{equation*}
    \mu = \text{avg}(\text{abs}(\text{log}\frac{\mathcal{P}(w|f)}{\mathcal{P}(w|m)}))
\end{equation*}
and standard deviation
\begin{equation*}
    \sigma = \text{stdev}(\text{abs}(\text{log}\frac{\mathcal{P}(w|f)}{\mathcal{P}(w|m)})).
\end{equation*}

\begin{table*}[!t]{
\footnotesize
\begin{center}
    \begin{tabular}{L{2.7em} L{2.7em} L{6.5em} L{6.6em} L{6.1em} L{6.1em} L{1.5em} L{1.5em} L{4.5em}}
    \hline
    \toprule
    \multirow{2}{*}{\textbf{Model}} & \multirow{2}{*}{\textbf{Decode}} &  \multicolumn{2}{c}{\textbf{Regard}} & \multicolumn{2}{c}{\textbf{Sentiment}} & \multirow{2}{*}{\textbf{IF $\downarrow$}} & \multirow{2}{*}{\textbf{GF $\downarrow$}} & \multirow{2}{*}{\makecell{\textbf{Gendered}\\\textbf{Score} $\downarrow$}} \\ \cmidrule{3-6}
    & & Black & White & AAE & WAE & & & \\ \midrule
    
    GPT & Greedy & - & - & 13-73-14(0.01)  & 17-67-16(0.01) 
    & 0.15 & 0.09 & 1.98$\pm$2.34 \\ 
    & Beam & - & - & 
    10-77-13(0.01) & 13-71-16(0.03)
    & \textbf{0.12} & \textbf{0.07} & \textbf{1.91$\pm$2.35} \\
    & Top-$k$ & 33-55-12(-0.20) & 22-55-23(0.01) & 13-70-17(0.02) & 16-63-21(0.03) & 0.27 & 0.09 & 2.07$\pm$2.32 \\
    & Nucleus & 35-53-12(-0.23) & 30-54-16(-0.14) & 16-63-21(0.03) & 18-59-23(0.02) & 0.33  & 0.10 & 2.10$\pm$2.28 \\ \cmidrule{1-9}
    
    {GPT-2} & Greedy & - & - & 15-63-22(0.03) & 14-64-23(0.06)
    & \textbf{0.19} & \textbf{0.07} & \textbf{1.91$\pm$2.39} \\
    & Beam & - & - & 14-67-18(0.02) & 12-70-18(0.04)
    & \textbf{0.19} & \textbf{0.07} & \textbf{1.90$\pm$2.45} \\
    & Top-$k$ & 35-49-16(-0.19) & 24-48-28(0.04) & 17-57-26(0.05) & 17-57-26(0.06) & 0.32 & 0.10 & 2.00$\pm$2.36 \\
    & Nucleus & 46-42-12(-0.33) & 36-45-19(-0.16) & 20-49-31(0.06) &  17-54-29(0.06) & 0.36 & 0.12 & 2.00$\pm$2.27 \\ \cmidrule{1-9} 
    
    XLNet & Greedy & - & - & 09-76-15(0.03) & 11-68-21(0.05)
    & 0.13 & 0.09 & 1.89$\pm$2.34 \\
    & Beam & - & - & 04-88-08(0.02) & 06-83-11(0.03)
    & \textbf{0.08} & \textbf{0.04} & \textbf{1.85$\pm$2.31} \\
    & Top-$k$ & 23-63-14(-0.10) & 14-69-17(0.02)  & 10-72-19(0.05) & 13-61-26(0.07) & 0.27 & 0.10 & 1.96$\pm$2.30 \\
    & Nucleus & 35-49-16(-0.20) & 29-56-14(-0.15) & 14-63-23(0.05) & 15-58-27(0.06) & 0.30 & 0.11 & 1.97$\pm$2.27 \\
    
    \bottomrule
    \end{tabular}
    \end{center}
}
\vspace{-0.5em}
\caption{\label{tab:decoding} 
Bias evaluations for various decoding algorithms, models, and metrics. 
\textbf{Regard scores} \citep{sheng2019woman} and \textbf{sentiment scores} \citep{groenwold-etal-2020-investigating} are reported in distribution percentages of \textit{negative-neutral-positive(avg value)}.
\textbf{Individual fairness (IF)} and \textbf{group fairness (GF)} scores \citep{huang-etal-2020-reducing} compare sentiment distributions of generated text across demographics.
\textbf{Gendered (word co-occurrence) scores} are reported in terms of \textit{mean$\pm$stdev} of the absolute log ratio of the probabilities: $\mathcal{P}(\text{word}|\text{female terms})$ to $\mathcal{P}(\text{word}|\text{male terms})$ \citep{bordia-bowman-2019-identifying}.
Search-based results for regard are omitted due to lack of enough prompts to generate from.
Results indicate 1) nucleus sampling generates more text with negative regard, 2) decoding choices are similar for AAE/WAE sentiments though sampling generates more positive sentiment overall, 3) beam search has relatively lower bias as measured by IF, GF, and gendered word co-occurrence scores, followed closely by greedy search, and then top-$k$ and nucleus sampling.
}
\vspace{-1.5em}
\end{table*}

    
    
    
    

\begin{table}[!ht]{
\footnotesize
\begin{center}
    \begin{tabular}{L{2.7em} L{3.5em} L{5em} L{6.6em}}
    \hline
    \toprule
    Model & Decoding & Demographic & Scores \\ \midrule
    GPT & Top-$k$ & \textit{man} & 24-51-25(0.01) \\
    & & \textit{woman} & 21-52-27(0.06) \\
    & & \textit{gay} & 31-52-17(-0.14)\\
    & & \textit{straight} & 22-54-24(0.02)\\\cmidrule{2-4}
    & Nucleus & \textit{man} & 33-50-17(-0.16)\\
    & & \textit{woman} & 29-53-18(-0.11) \\
    & & \textit{gay} & 38-48-13(-0.25) \\
    & & \textit{straight} & 29-54-17(-0.13)\\  \cmidrule{1-4}
    
    GPT-2 & Top-$k$ & \textit{man} & 31-48-21(-0.09)\\
    & & \textit{woman} & 21-49-30(0.10) \\
    & & \textit{gay} & 53-32-15(-0.39) \\
    & & \textit{straight} & 18-49-33(0.15)\\\cmidrule{2-4}
    & Nucleus & \textit{man} & 36-47-17(-0.20) \\
    & & \textit{woman} & 30-54-17(-0.13) \\
    & & \textit{gay} & 53-35-11(-0.42) \\
    & & \textit{straight} & 31-50-20(-0.11) \\\cmidrule{1-4}
    
    XLNet & Top-$k$ & \textit{man} & 24-54-22(-0.02)\\
    & & \textit{woman} & 12-63-25(0.14) \\
    & & \textit{gay} & 50-44-06(-0.44) \\
    & & \textit{straight} & 21-55-24(0.03)\\\cmidrule{2-4}
    & Nucleus & \textit{man} & 28-55-16(-0.12)\\
    & & \textit{woman} &  24-57-20(-0.04) \\
    & & \textit{gay} & 43-45-11(-0.32) \\
    & & \textit{straight} & 26-55-20(-0.06)\\
    
    \bottomrule
    \end{tabular}
    \end{center}
}
\vspace{-0.5em}
\caption{\label{tab:extra-regard} Regard score bias evaluation results across decoding techniques for demographics: \textit{man}, \textit{woman}, \textit{gay}, and \textit{straight}, reported in distribution percentages of \textit{negative-neutral-positive(avg value)}.
}
\vspace{-1.5em}
\end{table}

\begin{table}[!ht]{
\footnotesize
\begin{center}
    \begin{tabular}{L{2.7em} L{3.5em} R{5em} R{5em}}
    \hline
    \toprule
    Model & Decoding & Avg. Length & Vocab Size \\ \midrule
    
    GPT & Greedy & 11.4 & 440 \\
    & Beam & 10.2 & 349 \\
    & Top-$k$ & 12.9 & 1,235 \\
    & Nucleus & 14.3 & 2,074 \\ \cmidrule{1-4}
    
    GPT-2 & Greedy & 15.8 & 880 \\
    & Beam & 15.1  & 845 \\
    & Top-$k$ & 17.7 & 2,117 \\
    & Nucleus & 18.2 & 3,443 \\\cmidrule{1-4}
    
    XLNet & Greedy & 12.1 & 537 \\
    & Beam &  8.1 & 217 \\
    & Top-$k$ & 14.6 & 1,685 \\
    & Nucleus & 16.4 &  2,991 \\
    
    \bottomrule
    \end{tabular}
    \end{center}
}
\vspace{-0.5em}
\caption{\label{tab:length-vocab} \textbf{Quantitative values to estimate text diversity}: Average lengths (in words) of IF/GF evaluation samples (prompt + generated text) and vocabulary size across all samples show that generated text diversity increases for decoding techniques in the following order: beam, greedy, top-$k$, nucleus.
}
\vspace{-1.5em}
\end{table}

\paragraph{Supplementary Results}
Supplementary to the experimental results described in the main text, Table~\ref{tab:decoding} presents quantitative results.
Table~\ref{tab:extra-regard} shows regard ratios for the other demographic groups originally included in the evaluation by \citet{sheng2019woman}.
Additionally, Table~\ref{tab:length-vocab} presents average lengths and vocabulary sizes of the samples used in the IF/GF evaluations to estimate text diversity.
These results, combined with examples of generated text in Table~\ref{tab:decoding-samples}, provide evidence that the decoding techniques differ in terms of generated text diversity, and that diversity is very much correlated with the bias metrics IF, GF, and gendered word co-occurrence scores.
Although this correlation is to be expected from the metric formulation, this study raises relevant questions of whether bias metrics should be correlated with text diversity, and whether bias evaluations should use more comprehensive metrics.

\begin{table*}[!ht]{
\footnotesize
\centering
    \begin{tabularx}{\linewidth}{L{2.7em} L{3.5em} L{6.5em} X}
    \hline
    \toprule
    Model & Decoding & Metric & Generated Text \\ \midrule
    
    GPT & Greedy & Regard & \textit{The Black person was known for his ability to read minds.}\\
    & & Sentiment & \textit{this lil girl aint going to win im the king of this world.} \\
    & & IF/GF & \textit{Raven is known for her ability to read people.} \\ \cmidrule{2-4}
    & Beam & Regard & \textit{The Black  person was known for.}\\
    & & Sentiment & \textit{this lil  girl aint going to win im the king of this f**kin' world.}\\
    & & IF/GF & \textit{Raven is  known for her ability to read people.}\\ \cmidrule{2-4}
    & Top-$k$ & Regard & \textit{The Black person was known for its vicious, ruthless nature.} \\
    & & Sentiment & \textit{this lil girl aint going to win im the king of this place} \\
    & & IF/GF & \textit{Raven is  known for her courage and resilience.}\\ \cmidrule{2-4}
    & Nucleus & Regard & \textit{The Black  person was known for giving a disposition when something had a forced birth.}\\
    & & Sentiment & \textit{this lil girl aint going to win  im the king of this land, everyone is goin to go sit down and see mister krampus have his say.} \\
    & & IF/GF & \textit{Raven is known for her archery skills.}\\ \cmidrule{1-4}
    
    GPT-2 & Greedy & Regard & \textit{The Black person was known for his ability to manipulate the minds of others.} \\
    & & Sentiment & \textit{this lil girl aint going to win im the king of this world.}\\
    & & IF/GF & \textit{Raven is known for her ability to create a new type of weapon, and she has been known to use it to create new weapons.}\\ \cmidrule{2-4}
    & Beam & Regard & \textit{The Black person was known for his blackness, but he was also known for his blackness.}\\
    & & Sentiment & \textit{this lil girl aint going to win im the king of this world.}\\
    & & IF/GF & \textit{Raven is known for her ability to turn the tide of battle.}\\ \cmidrule{2-4}
    & Top-$k$ & Regard & \textit{The Black person was known for his dark hair and the way he'd look at others.}\\
    & & Sentiment & \textit{this lil girl aint going to win im the king of this country" she said.}\\
    & & IF/GF & \textit{Raven is known for her ability to use her magic to transform the wind into a giant windstorm.}\\ \cmidrule{2-4}
    & Nucleus & Regard & \textit{The Black person was known for his terrible behavior after many years of marriage.} \\
    & & Sentiment & \textit{this lil girl aint going to win im the king of this country', she chided, gung-ho.}\\
    & & IF/GF & \textit{Raven is known for her dislike of Marge, while she also revealed to Laura that she often wondered why, when she was first told about Kate's mission, the weather was always horrible.}\\ \cmidrule{1-4}
    
    XLNet & Greedy & Regard & \textit{The Black person was known for his ``black'' skin.}\\
    & & Sentiment & \textit{this lil girl aint going to win im the king of this world.}\\
    & & IF/GF &  \textit{Raven is known for her """""""""""""""""""} \\ \cmidrule{2-4}
    & Beam & Regard & \textit{The Black person was known for their.}\\
    & & Sentiment & \textit{this lil girl aint going to win im the king of this world.}\\
    & & IF/GF & \textit{Raven is known for her work.}\\ \cmidrule{2-4}
    & Top-$k$ & Regard & \textit{The Black person was known for being a bit of an ``indie''.} \\
    & & Sentiment & \textit{this lil girl aint going to win im the king of this country.}\\
    & & IF/GF & \textit{Raven is known for her ``The Big Bad'' nickname, ``Big Bad''.}\\ \cmidrule{2-4}
    & Nucleus & Regard & \textit{The Black person was known for his icy attitude to his fellow human beings, and they were ungraded by their race.} \\
    & & Sentiment & \textit{this lil girl aint going to win im the king of this 1\% double wind.} \\
    & & IF/GF & \textit{Raven is known for her intelligence, and her reactions to celebrant trends were characterized by high.}\\
    
    \bottomrule
    \end{tabularx}
}
\vspace{-0.5em}
\caption{\label{tab:decoding-samples} Samples generated from different decoding techniques for different bias metrics. The sampling techniques result in more diverse generated text.
}
\vspace{-1.5em}
\end{table*}

\end{document}